\documentclass[journal]{IEEEtran}

\usepackage{amsfonts}
\usepackage{setspace}

\usepackage{setspace}
\usepackage{amssymb}
\usepackage{epsfig}
\usepackage{graphicx}
\usepackage{xcolor}
\usepackage{algorithm}
\usepackage{algorithmicx}
\usepackage{algpseudocode}
\usepackage{amsmath}
\usepackage{cite}

\floatname{algorithm}{Algorithm}

\usepackage[ps2pdf,
bookmarks=false,
bookmarksnumbered=false, % true means bookmarks in
bookmarksopen=false, % true means only level 1
colorlinks=false]{}

\usepackage{tikz}
\newsavebox{\tempbox}

\begin{document}

\title{Robust Network Slicing: Multi-Agent Policies, Adversarial Attacks, and Defensive Strategies}

\author{\IEEEauthorblockN{Feng Wang,  M. Cenk Gursoy, and Senem Velipasalar}
\thanks{The authors are with the Department of Electrical Engineering and Computer Science, Syracuse University, Syracuse, NY, 13244
(e-mail: fwang26@syr.edu, mcgursoy@syr.edu, svelipas@syr.edu)}
\thanks{The material in this paper was presented in part at the 2022 IEEE International Conference on Communications (ICC).}
}

\maketitle

\begin{abstract}
In this paper, we present a multi-agent deep reinforcement learning (deep RL) framework for network slicing in a dynamic environment with multiple base stations and multiple users. %, a jammer, and the policy ensemble for both agents in this competitive environment. 
%We first introduce the wireless network virtualization (WNV) and the interference channel model, formulate the network slicing problem in the dynamic environment in which fading varies, users have mobility, and requests are randomly generated over time. 
In particular, we propose a novel deep RL framework with multiple actors and centralized critic (MACC) in which actors are implemented as pointer networks to fit the varying dimension of input. We evaluate the performance of the proposed deep RL algorithm via simulations to demonstrate its effectiveness. 
Subsequently, we develop a deep RL based jammer with limited prior information and limited power budget. The goal of the jammer is to minimize the transmission rates achieved with network slicing and thus degrade the network slicing agents' performance. We design a jammer with both listening and jamming phases and address jamming location optimization as well as jamming channel optimization via deep RL. We evaluate the jammer at the optimized location, generating interference attacks in the optimized set of channels by switching between the jamming phase and listening phase. We show that the proposed jammer can significantly reduce the victims' performance without direct feedback or prior knowledge on the network slicing policies. Finally, we devise a Nash-equilibrium-supervised policy ensemble mixed strategy profile for network slicing (as a defensive measure) and jamming. We evaluate the performance of the proposed policy ensemble algorithm by applying on the network slicing agents and the jammer agent in simulations to show its effectiveness. %with or without the existence of an adversary.

\end{abstract}


\section{Introduction}
Network slicing in 5G radio access networks (RANs) allows enhancements in  service flexibility for novel applications with heterogeneous requirements \cite{ericsson20175g, foukas2017network, alliance2016description, zhang2017network}. In network slicing, the physical cellular network resources are divided into multiple virtual network slices to serve the end user, and thus network slicing is a vital technology to meet the strict requirements of each user by allocating the desired subset of network slices \cite{kazmi2019network}. Instead of model-based optimization of resource allocation  that assumes the knowledge of traffic statistics \cite{ye2018dynamic, peng2019spectrum, tang2019service}, deep reinforcement learning (deep RL) \cite{ye2019deep, xu2017deep} as a model-free decision making strategy can be deployed to optimize slice selection and better cope with the challenges such as dynamic environment, resource interplay, and user mobility \cite{li2018deep, zhang2019deep, shi2020reinforcement, shao2021graph}. Most existing studies in the literature assume the slice state to be identical for all different slices over time, and hence the selection problem reduces to assigning the number of slices to each request. 

In this paper, we consider a more practical scenario of a cellular coverage area with multiple base stations and a dynamic interference environment, analyze resource allocation in the presence of multiple users with random mobility patterns, and develop a multi-agent actor-critic deep RL agent to learn the slice conditions and achieve cooperation among base stations. We %{in \cite{wang2022multi}} 
propose a learning and decision-making framework with multiple actors and a centralized critic (MACC) %\cite{lyu2021contrasting, foerster2018counterfactual, lowe2017multi} 
that aims at maximizing the overall performance instead of local performance. 
{In the machine learning literature, MACC framework has been proposed for general RL tasks \cite{lyu2021contrasting, foerster2018counterfactual, lowe2017multi}.}
In our setting, this multi-agent system requires the actors at each base station to communicate with a centralized critic at the data center/server to share the experience and update parameters during the training process. It subsequently switches to decentralized decision making following the training. 

Typically, when 5G RAN achieves faster transmission rates with higher frequency bands, it also has smaller coverage, necessitating a relatively dense cellular network architecture. In such a setting, the dynamic environment with user mobility might have significant impact on the network slicing performance. Due to this, we introduce the pointer network \cite{vinyals2015pointer} to implement the actor policy to handle the varying observations of the deep RL agents.% We compare this proposed algorithm with different combinations of decentralized critic, feed-forward neural network (FNN) based actor, and other statistical algorithms to show the outstanding performance of the proposed novel approach. 

It is important to note that due to being highly data driven, deep neural networks are vulnerable to minor but carefully crafted perturbations, and it is known that adversarial samples with such perturbations can cause significant loss in accuracy, e.g. in  inference and classification problems in computer vision \cite{goodfellow2014explaining, dong2018boosting, lu2020enhancing}. Given the broadcast nature of wireless communications,  deep learning based adversarial attacks have also recently been attracting increasing attention in the context of wireless security \cite{shi2018adversarial, wang2021resilient}. {In particular, deep RL agents are vulnerable to attacks that lead to adversarial perturbations in their observations (akin to adversarial samples in classification and inference problems). In wireless settings, jamming is an adversarial attack that alters the state and/or observations of decision-making RL agents.} Motivated by these considerations, we also  design a deep RL based jammer agent with jamming and listening phases. Different from most existing works, we analyze how the jamming location and channel selection are optimized without direct feedback on the victims' performance. We further analyze the performance degradation in network slicing agents in the presence of jamming attacks and identify their sensitivity in an adversarial environment.  

%It is important to note that deep neural networks (due to being highly data-driven) are vulnerable to the worst-case perturbation of minor scale but being carefully crafted, and it is known that adversarial samples with such perturbation will cause significant loss to the classification problem in computer vision \cite{goodfellow2014explaining, dong2018boosting, lu2020enhancing}. Due to the broadcast nature of wireless communication, the deep learning based adversarial attack also leads to increasing attention to wireless security \cite{shi2018adversarial, wang2021resilient}. Motivated by these considerations, we in this paper consider the aforementioned WNV system and MACC network slicing agent, and design a deep RL based jammer agent with jamming phase and listening phase. Different from most existing works, we analyse how the jamming location and channel selection are optimized without direct feedback corresponding to the victims' performance.

Finally, we note that there is also a growing interest in developing defense strategies against adversarial attacks \cite{madry2017towards, zhang2019theoretically, dong2019evading}, and specifically jamming attacks \cite{shi2021attack, wang2021adversarial}. One of the most intriguing defense strategies is to ensemble several different policies, explore alternative strategies, and provide stable performance \cite{lowe2017multi, khadka2019collaborative}. In our context, we consider the mobile virtual network operator (MVNO) and the jammer as two players in a zero-sum incomplete information game. Several existing studies within this framework focus on applications, e.g., involving  video games or poker games, in which random exploration does not lead to physical loss or penalty. In contrast, wireless users may experience  disconnectivity in such cases. Therefore, an efficient and prudent exploration plan with fast convergence is desired. Most existing works also restrict the scope to the performance of the considered player or its performance against a certain type of adversary, failing to consider the nature of the zero-sum game against an unknown opponent that is potentially adaptive. In this paper, we propose an approach we refer to as Nash-equilibrium-supervised policy ensemble (NesPE) that utilizes the optimized mixed strategy profile to supervise the training process of the policies in the ensemble to fully explore the environment, and leaves no improvement space for the opponent to pursue the global equilibrium over all possible policy ensembles. We evaluate the performance of NesPE by applying it on both the aforementioned network slicing victim agent and jammer agent in a competitive context,  and compare its significantly improved performance with two other policy ensemble algorithms. 

The remainder of the paper is organized as follows. In Section \ref{sec: sys}, we introduce the wireless network virtualization framework and dynamic channel model, and formulate the network slicing problem. In Section \ref{sec: ac}, we propose the MACC deep RL algorithm with its pointer network architecture for actor implementation, and describe the network slicing agents. In this section, we also evaluate the performance of the proposed algorithm.
%, and provide comparisons with other statistical and deep RL based algorithms. 
Subsequently, we devise the deep RL based jammer agent in Section \ref{sec: jam}, introduce the two operation phases, jamming location optimization, jamming channel optimization and the actor-critic implementation, and evaluate the performance %to show the effectiveness of proposed listening measurement based on the impact factor $\beta$. 
Finally, we design the NesPE algorithm in Section \ref{sec: equil}, describe the steps of the algorithm, analyze its performance in both non-competitive and competitive environments, and compare it with other policy ensemble algorithms by applying on both the victim agent and jammer agent.
Finally, we conclude the paper in Section \ref{sec: con}.

% \section{Multi-Agent Reinforcement learning for Network Slicing}\label{sec:user}
% Rewrite the article before jammer section into one introduction section!
% cite published paper \cite{wang2022multi}

\section{System Model and Problem Formulation}\label{sec: sys}
In this section, we introduce the wireless network virtualization (WNV) and the interference channel model, and formulate the network slicing as an optimization problem.

\subsection{Wireless Network Virtualization}\label{subsec: wnv}
\begin{figure}
	\centering
	\includegraphics[width=1\linewidth]{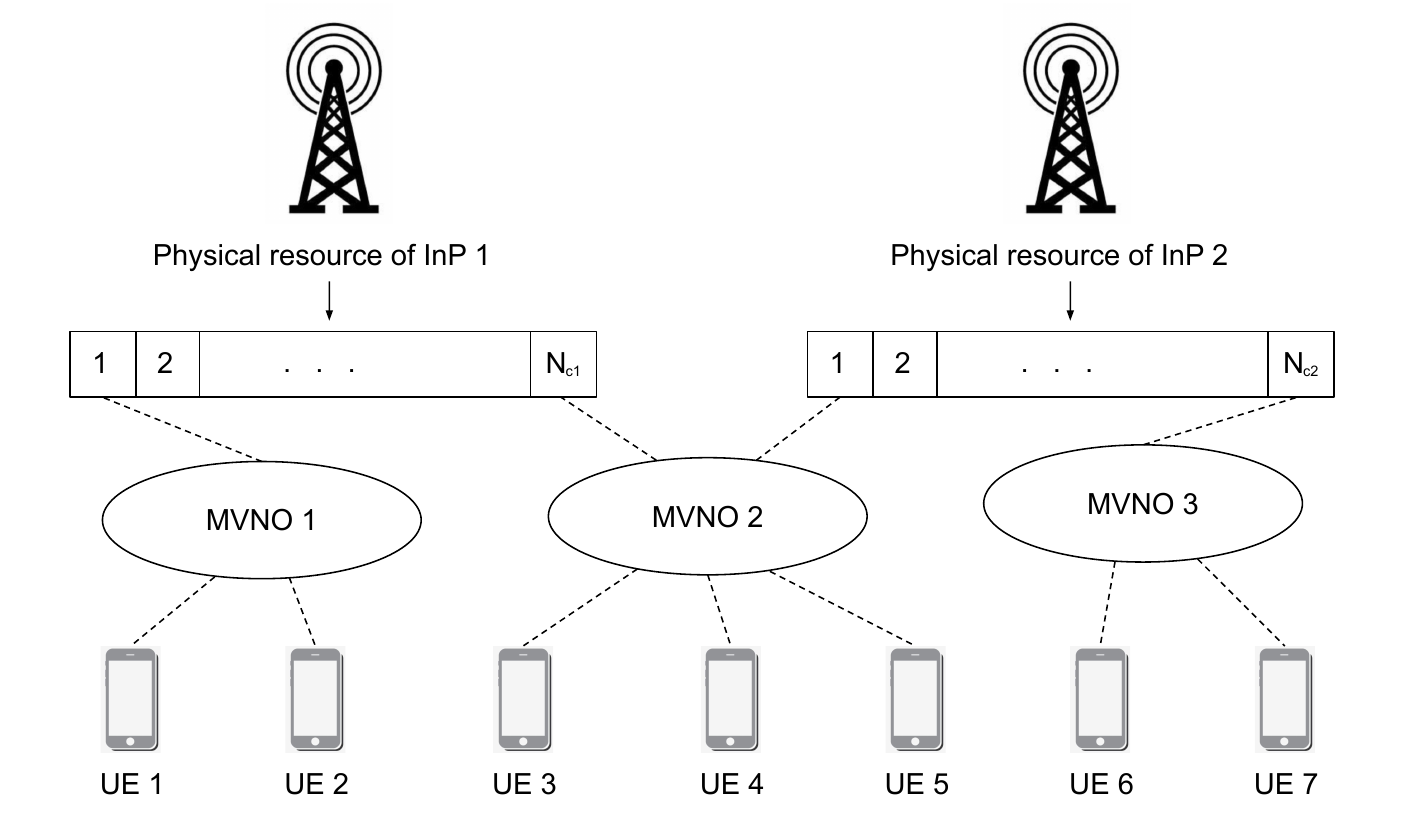}
	\caption{WNV system model for allocating virtualized physical resources via MVNO. {$N_{cb}$ denotes the number of slices available at InP $b$.}}
	\label{fig:wnv_sys}
\end{figure}

WNV is well-known for enhancing the spectrum allocation efficiency. As shown in Fig. \ref{fig:wnv_sys}, infrastructure providers (InPs) own the physical infrastructure (e.g., base stations, backhaul, and core network), and operate the physical wireless network. WNV virtualizes the physical resources of InPs and separates them into slices to efficiently allocate these virtualized resources to the mobile virtual network operators (MVNOs). Therefore, MVNOs deliver differentiated services via slices to user equipments (UEs) with varying transmission requirements. 
%Thus, MVNO removes the tight coupling between the cellular infrastructure and mobile networks through WNV and adds more flexibility to resource allocation. In Fig. \ref{fig:bs_sys}, each MVNO may be served by a group of InPs (or a group of base stations of the same InP) in the service area, and MVNO may share the purchased resource to multiple users within the corresponding base station coverage.

In this paper, we consider a service area that contains $N_B$ base stations of InPs with inter-base-station interference, and MVNOs that require services from the InPs. These MVNOs provide services to $N_u$ UEs that are within the coverage area of at least one base station, and may move from the coverage of one base station to that of another. Users in the coverage area of multiple base stations that serve the MVNO can be assigned to any of these base stations. Without loss of generality, we discuss the station-user pair for a single MVNO in the remainder of this paper. Next, we introduce the dynamic channel environment.

%\begin{figure}
%	\centering
%	\includegraphics[width=1.0\linewidth]{BS system model.pdf}
%	\caption{System model of multi-InP resource allocation. }
%	\label{fig:bs_sys}
%\end{figure}

\subsection{Interference Channel Model}\label{subsec: ch}
We consider a dynamic environment where $N$ channels are available for each MVNO. The fading coefficient of the link between base station $b$ and a UE $u$ in a certain channel $c$ at time $t$ is denoted by $h_{c}^{b,u}(t) \sim \mathcal{CN} (0,1)$, an independent and identically distributed (i.i.d.) circularly symmetric complex Gaussian (CSCG) random variable with zero mean and unit variance. Each fading coefficient varies every $T$ time slots as a first-order complex Gauss-Markov process, according to the Jakes fading model \cite{liang2017spectrum}. Therefore, at time $(n+1)T$, the fading coefficient can be expressed as
\begin{equation} \label{eq:fading}
h_{c}^{b,u}((n+1)T) = \rho h_{c}^{b,u}(nT) + \sqrt{1-\rho^2} e_{c}((n+1)T),
\end{equation}
where $e_{c}$ denotes the channel innovation process, which is also an i.i.d. CSCG random variable. Furthermore, we have $\rho=J_0(2\pi f_d T)$, where $J_0$ is the zeroth order Bessel function of the first kind and $f_d$ is the maximum Doppler frequency.

\subsection{Network Slicing Problem Formulation}\label{subsec: ns}
In the aforementioned environment, the resource allocation is performed in a first-come first-served fashion. Each base station may serve at most $N_r$ requests from different users simultaneously, and the request queue can stack at most $N_q$ requests. For each request, multiple channels can be allocated, and the transmission power is $P_B$ in each channel. Different requests served by the same base station do not share the same channels, while requests to different base stations may share the same channel at the cost of inflicting interference. For all requests to each base station, transmission is allowed in no more than $N_c$ channels, and consequently the total power is limited to $N_c P_B$. In this setting, if request $k$ is allocated the subset $C_k$ of channels at base station $b$, the sum rate $\mathsf{r}_k$ for this request is
\begin{equation} \label{eq:sum_rate}
\mathsf{r}_k = \sum_{c \in C_k} {\mathsf{r}_c^{b,u} },
\end{equation}
with
\begin{equation} \label{eq:individual_rate}
\mathsf{r}_c^{b,u} = \log_2 \left( 1+{\frac{P_B L^{b,u} |h_{c}^{b,u}|^2}{\sum_{b' \neq b}{\mathcal{N}^{b, b'}_c} + \sigma^2}} \right) ,
\end{equation}
\begin{equation} \label{eq:user_intf}
\mathcal{N}^{b, b'}_c = \mathbf{1}_c^{b,b'} P_B L^{b',u} |h_{c}^{b',u}|^2 ,
\end{equation}
where $c$ denotes the index of a selected channel, $\mathsf{r}_c^{b,u}$ is the transmission rate from base station $b$ to UE $u$ in channel $c$, $\mathcal{N}^{b, b'}_c$ is the interference from base station $b'$ in channel $c$, $\mathbf{1}_c^{b,b'}$ is the indicator function for transmission at base stations $b$ and $b'$ sharing channel $c$, $\sigma^2$ is the variance of the additive white Gaussian noise, and $L^{b,u}$ is the path loss:
\begin{equation} \label{eq:pathloss}
\begin{aligned} 
L^{b,u} & = \left(h_B^2 + (x_b - x_u)^2 + (y_b - y_u)^2\right)^{\alpha/2},
\end{aligned}
\end{equation}
where $h_B$ is the height of each base station, $\alpha$ is the path loss exponent, and $\{x_b, y_b\}$ and $\{x_u, y_u\}$ are the 2-D coordinates of base station $b$ and user $u$, respectively. 

Therefore, for each base station, the resource assignment including the selected subsets $C_k$, the number of selected channels $n_c$, the number of served requests $n_r$, and the number of requests in queue $n_q$ must follow the following constraints:
\begin{equation} \label{eq:const00}
{C_{k_1}}\cap{C_{k_2}} = \text{\O}, \forall 1 \le k_1 < k_2 \le n_r , 
\end{equation}
\begin{equation} \label{eq:const01}
n_c = \left| \cup_{k=1}^{n_r}{C_k} \right| ,
\end{equation}
\begin{equation} \label{eq:const02}
n_c \le N_c ,
\end{equation}
\begin{equation} \label{eq:const03}
n_r \le N_r ,
\end{equation}
\begin{equation} \label{eq:const04}
n_q \le N_q .
\end{equation}

{Above, (\ref{eq:const00}) indicates that no channel is shared among different requests at the same base station, and (\ref{eq:const01}) defines the number of channels being used. The constraints in (\ref{eq:const02})--(\ref{eq:const04}) are the upper bounds on the number of channels, the number of requests being served simultaneously, and the number of requests in the queue, respectively.} If $n_r = N_r$ and $n_q = N_q$ at a base station, any incoming request to that base station will be denied service. 

The features of each request $k$ consist of minimum transmission rate $m_k$, lifetime $l_k$, and initial payload $p_k$ before the transmission starts. At each time slot $t$ during transmission, the constraints are given as follows:
\begin{equation} \label{eq:const1}
\mathsf{r}_k(t) \ge m_k,
\end{equation}
\begin{equation} \label{eq:const2}
l_k(t) > 0, 
\end{equation}
where $l_k(t)$ denotes the remaining lifetime at time $t$. Note that with this definition, we have $l_k(0) = l_k$. If any request being processed fails to meet the constraints (\ref{eq:const1}) or (\ref{eq:const2}), it will be terminated and be marked as failed. Any request in the queue that fails to meet constraint (\ref{eq:const2}) will also be removed and marked as failure. Otherwise, the lifetime will be updated for each request being processed and each request in the queue as follows:
\begin{equation} \label{eq:update2}
l_k(t+1) = l_k(t) - 1.
\end{equation}
Each request being processed will transmit $\mathsf{r}_k(t)$ bits of the remaining payload:
\begin{equation} \label{eq:update1}
p_k(t+1) = \operatorname*{max} (p_k(t) - \mathsf{r}_k(t), 0).
\end{equation}
Note that the initial payload is $p_k(0) = p_k$. If the payload is completed within the lifetime (i.e., the remaining payload is $p_k(t+1) = 0$ for $t+1 \le l_k$), the request $k$ is completed and marked as success.

For all aforementioned cases, if the request is completed successfully in time, the network slicing agent at the base station $b$ receives a positive reward $R_k$ equal to the initial payload $p_k$. Otherwise, it receives a negative reward of $R_k = -p_k$. Each user can only send one request at a time. 

Afterwards, base station $b$ records the latest transmission rate history of $\mathsf{r}_c^{b,u}$ into a 2 dimensional matrix $H^b \in \mathbb{R}_{\ge 0}^{N_u \times N}$. In each time slot $t$, for request $k$ from user $u$ that is allocated subset $C_k$ of channels at base station $b$, the base station updates the entries corresponding to the channels that were selected for transmission:
\begin{equation} \label{eq:updateH}
H^b[u, c] = \mathsf{r}_c^{b,u}(t), \forall c \in C_k.
\end{equation}

Note that the location $\{x_u, y_u\}$ of user $u$, the fading coefficient $h_{c}^{b,u}$, and the interference corresponding to $\mathbf{1}_c^{b,b'}$ vary over time, and therefore $H_b$ is only a first-order approximation of the potential transmission rate $\mathsf{r}_c^{b,u}(t+1)$. 

The goal of the network slicing agent at each base station is to find a policy $\pi$ that selects $n_c$ channels and assigns them to $n_r$ requests, so that the sum reward of all requests at all base stations is maximized over time:
\begin{equation} \label{eq:sumrateopt}
\operatorname*{argmax}_{\pi}{ \sum_{t'=t}^{\infty}{ \left(\gamma^{(t'-t)} \sum_{b=1}^{N_B}\sum_{k \in K'_b}{R_k} \right) } },
\end{equation}
where $K'_b$ is the set of completed or terminated requests for base station $b$ at time $t$, and $\gamma \in (0, 1)$ is the discount factor.

\section{Multi-Agent Deep RL with Multiple Actors and Centralized Critic (MACC)}\label{sec: ac}
To solve the problem in (\ref{eq:sumrateopt}), we propose a multi-agent deep RL algorithm with multiple actors and centralized critic (MACC) where the actors (one at each base station) utilize the pointer network to achieve the goal of attaining the maximal reward over all base stations by choosing the optimal subset of channels in processing each request. In the remainder of this paper, we denote the full observation at base station $b$ as $\mathcal{O}_b$, the observation over all base stations as $\mathcal{O} = \cup_{b=1}^{N_B} \mathcal{O}_b$, and the channel selection at base station $b$ as a matrix of actions $\mathcal{A}_b \in \{0, 1\}^{n_r \times N}$. Each element in $\mathcal{A}_b$ is an indicator: 
\begin{align}\label{eq:indi_act}
\mathcal{A}_b[k, c] = \begin{cases}
1 \hspace{.3cm} \text{if channel $c$ is assigned to request $k$,}\\
0 \hspace{.3cm} \text{otherwise.}
\end{cases}
\end{align}
In each time slot when $n_c(t)$ channels are selected, $\sum_{c}\sum_{k}\mathcal{A}_b[k, c] = n_c(t)$. In this section, we first introduce the MACC framework, then describe the pointer network as the actor structure, and finally analyze the implementation on the considered network slicing problem.

\subsection{Deep RL with Multiple Actors and Centralized Critic}\label{subsec: macc}
In this section, we briefly discuss the actor-critic algorithm \cite{peters2008natural}. This deep RL algorithm utilizes two neural networks, the actor and critic. The two networks have separate neurons and utilize separate backpropagation, and they may have separate hyperparameters. 

The multi-agent extension of actor-critic algorithm where each agent has separate actor and critic that aim at maximizing each individual reward is referred to as independent actor-critic (IAC) \cite{lyu2021contrasting, foerster2018counterfactual}. 
%For the agent that chooses channels at base station $b$, the actor with parameter $\theta$ and policy $f^\theta(\mathcal{O}_b)$ maps the input local observation $\mathcal{O}_b$ to the output action $\mathcal{A}_b$, which is similar to a Q-value generator. The critic with parameter $\phi'$ and policy $g^{\phi'}(\mathcal{O}_b)$ maps $\mathcal{O}_b$ to a single temporal difference (TD) error:
%\begin{equation} \label{eq:acdelta_icc}
%\delta_b(t) = \sum_{k \in K_b}{R_k}(t) + \gamma %g^{\phi'}(\mathcal{O}_b(t)) - %g^{\phi'}(\mathcal{O}_b(t-1)),
%\end{equation}
%where $K_b$ is the set of requests being processed and requests in the queue at time $t$, and $\gamma \in (0, 1)$ is the discount factor. During each training period, the critic is updated towards achieving the optimized parameter $\phi'^*$ to minimize the least square temporal difference:
%\begin{equation} \label{eq:lstd_icc}
%\phi'^*=\operatorname*{argmin}_{\phi'}{(\delta_b^{g_{\phi'}})^2}.
%\end{equation}
%The actor is updated towards the optimized parameter $\theta^*$ to minimize the policy gradient:
%\begin{equation} \label{eq:pg_icc}
%\theta^*=\operatorname*{argmax}_{\theta}{\nabla_\theta \log f^\theta (\mathcal{O}_b) \delta_b^{g_{\phi'}}} .
%\end{equation}
%Both networks are updated alternately to attain the optimal actor-critic policy.
When the action of each agent interferes with the others and the goal is to maximize the sum reward over all agents, the deep RL with MACC may be preferred \cite{foerster2018counterfactual}. In this framework, the decentralized actors with parameter $\theta$ and policy $f^\theta(\mathcal{O}_b)$ are the same as in IAC, while there is only one centralized critic with parameter $\phi$ and policy $g^\phi(\mathcal{O})$ aiming at the sum reward by updating each decentralized actor. Therefore, MACC agents are more likely to learn coordinated strategies through interaction between agents and mutual environment, despite the scarcity of information sharing at the training phase. %Instead of (\ref{eq:acdelta_icc}), (\ref{eq:lstd_icc}), and (\ref{eq:pg_icc}), 
In the framework of MACC, the critic maps $\mathcal{O}$ to a single temporal difference (TD) error
\begin{equation} \label{eq:acdelta_macc}
\small
\delta(t) = \sum_{b=1}^{N_B}\sum_{k \in K_b}{R_k}(t) + \gamma g^\phi \left(\mathcal{O}(t) \right) - g^\phi \left(\mathcal{O}(t-1) \right),
\end{equation}
\normalsize
where $K_b$ is the set of requests being processed and requests in the queue at time $t$, and $\gamma \in (0, 1)$ is the discount factor. Then, the critic parameters $\phi$ and actor parameters $\theta$ are optimized by considering the least square temporal difference and policy gradient, respectively, as follows:    
\begin{equation} \label{eq:lstd_macc}
\phi^*=\operatorname*{argmin}_{\phi}{(\delta^{g_\phi})^2},
\end{equation}
\begin{equation} \label{eq:pg_macc}
\theta^*=\operatorname*{argmax}_{\theta}{\nabla_\theta \log f^\theta (\mathcal{O}_b) \delta^{g_\phi}} .
\end{equation}

For a wireless resource allocation problem, fast convergence in the training of neural networks is highly important. Therefore in our implementation, we not only recommend offline pre-training via simulations, but also speed up learning by sharing the neural network parameters among the agents, i.e., we use all the information to update one actor and one critic (as in IAC) during training and share the parameters among all agents. %Given different local observations, the corresponding local actors will always behave in an optimized way. Compared to distributed training only with local data, this training method with data from multiple agents also avoids over-fitting to a certain local environment.

%During the online training of MACC agents, all the information of $\mathcal{O}$ is required to obtain the TD error and update with respect to every action-reward pairs. Therefore with a period of $T_t$, all agents upload local observations to the data center for training, and the center distributes the update of actor parameters $\Delta_{\theta} = \theta(t+1) - \theta(t)$ (which generally requires less bits of data compared to $\theta(t+1)$) to each agent. Note that during each parameter update for both IAC and MACC, the mini-batch of training samples (i.e., action-reward pairs) should be randomly drawn from a longer history record for faster convergence.  

\subsection{Pointer Network}\label{subsec: ptr}
In this section, we introduce the pointer network to implement the actor policy $f^\theta$ for IAC and MACC. 

Traditional sequence-processing recurrent neural networks (RNN), such as sequence-to-sequence learning \cite{sutskever2014sequence} and neural Turing machines \cite{graves2014neural}, have a limit on the output dimensionality, i.e., the output dimensionality is fixed by the dimensionality of the problem so it is the same during training and inference. However in a network slicing problem, the actor's input dimensionality $\{N, n_c, n_r\}$ may vary over time, and thus the expected output dimension of $\mathcal{A}_b$ also varies according to the input. As opposed to the aforementioned sequence-processing algorithms, pointer network learns the conditional probability of an output sequence with elements that are discrete tokens corresponding to positions in the input sequence \cite{vinyals2015pointer}, and therefore the dimension of action $\mathcal{A}_b$ in each time slot depends on the length of the input.

Another benefit of pointer networks is that they reduce the cardinality of $\mathcal{A}_b$. General deep RL algorithms typically list out each combination as an element in action $\mathcal{A}_b'$, and each element indicates picking $n_c$ channels out of $N$ channels and assigning each to one of $n_r$ requests. Therefore $\mathcal{A}_b'$ has the dimension of $\binom{N}{n_c} n_r^{n_c}$. When the dimensions of $N$, $n_c$ and $n_r$ are high, this will require a prohibitively large network to give such an output, and require exponentially longer time to train. Comparatively, $\mathcal{A}_b$ generated from our proposed pointer network actor has only $n_r N$ output elements. 

\begin{figure}
	\centering
	\includegraphics[width=0.8\linewidth]{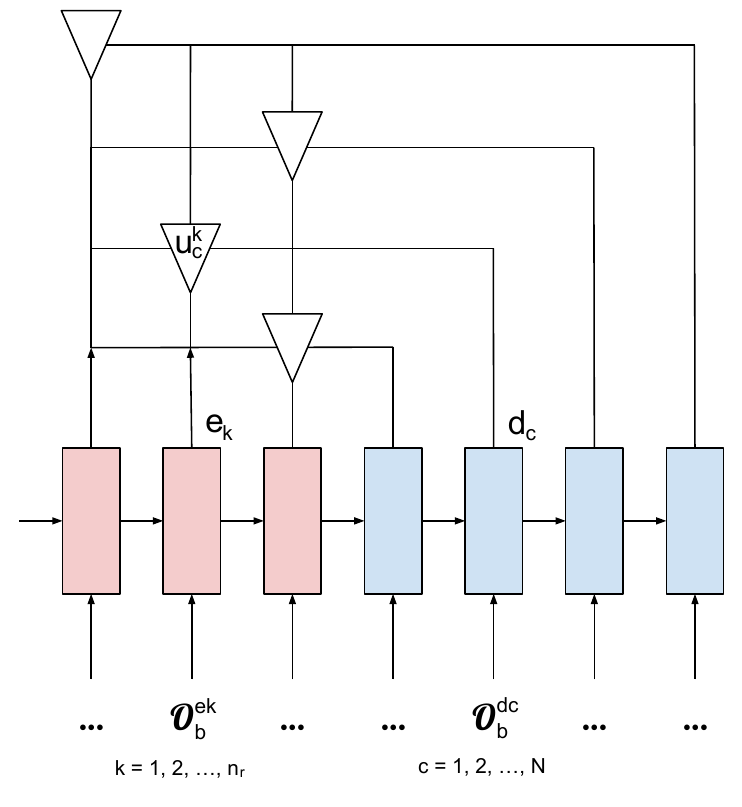}
	\caption{Pointer network structure. At each step, the encoder RNN (red) converts each element of input $\mathcal{O}_b^{ek}$ into output $e_k$, while the hidden state and cell state are fed to the decoder RNN (blue). The decoder converts each element of input $\mathcal{O}_b^{dc}$ into output $d_c$ at each step. The final output of the pointer network is generated by attention mechanism as the conditional probability $\mathcal{P}^c$, and each element are generated via $e_k$ and $d_c$.}
	\label{fig:ptr}
\end{figure}

Therefore, we here introduce the pointer network as the novel architecture of the actor of the proposed agents as shown in Fig. \ref{fig:ptr}. Similar to sequence-to-sequence learning, pointer network utilizes two RNNs called encoder and decoder whose output dimensions are $h_e$ and $h_d$. The former encodes the sequence $\{\mathcal{O}_b^{e1}, \mathcal{O}_b^{e2}, \ldots, \mathcal{O}_b^{e{n_r}}\}$ with $\mathcal{O}_b^{ek} \subset \mathcal{O}_b$ (where the index $k \in \{ 1, 2, \ldots, n_r\}$ corresponds to requests), and produces the sequence of vectors $\{e_1, e_2, \ldots, e_{n_r}\}$ with $e_k \in \mathbb{R}^{h_e}$ at the output gate in each step. The decoder inherits the encoder's hidden state and is fed by another sequence $\{\mathcal{O}_b^{d1}, \mathcal{O}_b^{d2}, \ldots, \mathcal{O}_b^{d{N}}\}, \mathcal{O}_b^{dc} \subset \mathcal{O}_b$ (where the index $c \in \{1, 2, \ldots, N \}$ corresponds to channels), and produces the sequence of vectors $\{d_1, d_2, \ldots, d_{N}\}, d_c \in \mathbb{R}^{h_d}$. 

Furthermore, pointer network computes the conditional probability array $\mathcal{P}^c \in [0, 1]^{n_r}$ where
\begin{equation} \label{eq:p_ptr}
\mathcal{P}^c[k] = P(\mathcal{A}_b[k, c] = 1 | \mathcal{O}_b^{d1}, \ldots, \mathcal{O}_b^{d{c}}, \mathcal{O}_b^{e1}, \ldots, \mathcal{O}_b^{e{n_r}}),
\end{equation}
using an attention mechanism as follows:
\begin{equation} \label{eq:att_ptr}
\begin{aligned} 
u^c_k & = v^T \operatorname*{tanh} (W_1 e_k + W_2 d_c) ,  \\
\mathcal{P}^c & = \operatorname*{softmax} ([u^c_1, u^c_2, \ldots, u^c_{n_r}]^T),
\end{aligned} 
\end{equation}
where softmax normalizes the vector of $u^c_k$ to be an output distribution over the dictionary of inputs, and vector $v$ and matrices $W_1$ and $W_2$ are trainable parameters of the attention model.

We note that the $n_r \times N$ matrix of the pointer network output $\mathcal{P} = [\mathcal{P}^1, \mathcal{P}^2, \ldots, \mathcal{P}^N]$ typically consists of non-integer values, and we obtain the decision of $\mathcal{A}_b$ by picking $n_c$ maximum elements in each column:
\begin{equation} \label{eq:p_to_act}
\begin{aligned} 
M_1 & = [\operatorname*{max}(\mathcal{P}^1), \operatorname*{max}(\mathcal{P}^2), \ldots, \operatorname*{max}(\mathcal{P}^N)] , \\
M_2 & = \operatorname*{argmax}_{M_1' \subset M_1, |M_1'|=n_c} {\sum_{m \in M_1'}{m}} , \\
\mathcal{A}_b[k, c] & = \begin{cases}
1 \hspace{.3cm} \text{if $c \in M_2$ and $k = \operatorname*{argmax}\mathcal{P}^c$,}\\
0 \hspace{.3cm} \text{otherwise.}
\end{cases}\\
\end{aligned} 
\end{equation}

In our implementation, we use two Long Short Term Memory (LSTM) \cite{hochreiter1997long} RNNs of the same size (i.e., $h_e = h_d$) to model the encoder and decoder. During the backpropagation phase of neural network training, the partial derivative of error functions with respect to weights in layers (encoder and decoder in our case) that are far from the output layer may be vanishingly small, preventing the network from further training, and this is called vanishing gradient problem \cite{pascanu2013difficulty}. However, the encoder and decoder are cardinal for the function of the pointer network while we demand fast convergence, so we set $W_1$ and $W_2$ in (\ref{eq:att_ptr}) as trainable vectors and set $v$ as a single constant to reduce the depth of the pointer network and speed the training on both LSTMs.

\subsection{Network Slicing Agent}\label{subsec: agent}
In this subsection, we introduce the action and state spaces for MACC deep RL with pointer networks employed for the aforementioned network slicing problem. 

\subsubsection{Centralized Critic}\label{subsubsec: critic}
The centralized critic agent $g^\phi(\mathcal{O})$ is implemented as a feed-forward neural network (FNN) that uses ReLU activation function for each hidden layer. This FNN takes the observation $\mathcal{O} = \{\mathcal{O}_1, \mathcal{O}_2, \ldots, \mathcal{O}_{N_B}\}$ as the input where $\mathcal{O}_b(t) = \{\mathcal{A}_b(t-1), H^b[\mathcal{U}_b(t), \mathcal{C}](t), \mathcal{I}_b(t)\}$. $\mathcal{A}_b(t-1)$ is the action matrix as described in (\ref{eq:p_to_act}) in the last time slot, $H^b(t)$ is the transmission rate history described in (\ref{eq:updateH}), $\mathcal{U}_b(t)$ is the set of users corresponding to the $n_r$ requests served by base station $b$, and $\mathcal{C}$ is the set of all channels $\{1, 2, \ldots, N\}$. $\mathcal{I}_b$ is the information of requests, including the remaining payload $p_k$, minimum rate $m_k$, remaining lifetime $l_k$ and the absolute value of reward $|R|=p_k$ of the requests being processed and the requests in the queue: $\mathcal{I}_b=\mathcal{I}_b^{K_b}$ where $\mathcal{I}_b^k = \{p_k, m_k, l_k, p_k\}$.

The output of $g^\phi(\mathcal{O})$ is a single value that aims at minimizing the error $\delta(t)$ in (\ref{eq:acdelta_macc}). Note that in our implementation, the sum reward is not the instantaneous reward at time $t$, but the rewards of $K_b$, the set of all requests being processed and requests in the queue at time $t$. Therefore, we do not get the reward value or use the sample for training until all requests and those in the queue are completed. Compared to the instantaneous reward, we use this reward assignment to strengthen the correlation between the action $\mathcal{A}_b$ and the outcome, and thus speed the convergence.

\subsubsection{Decentralized Actor with Pointer Network}\label{subsubsec: actor}
The decentralized actor agent $f^\theta(\mathcal{O}_b)$ at each base station $b$ is implemented as a pointer network. The initial hidden state and cell state of the encoder LSTM are generated by a separate FNN without hidden layers, whose input is the information of the requests in the queue $\{\mathcal{I}_b^{n_r+1}, \mathcal{I}_b^{n_r+2}, \ldots, \mathcal{I}_b^{n_r+n_q}\}$, and the output is a vector of two states with length $2 h_e$. 

The input that the encoder LSTM receives at each step is $\mathcal{O}_b^{ek}(t) = \{\mathcal{A}_b[k, \mathcal{C}](t-1), H^b[\mathcal{U}_b[k](t), \mathcal{C}](t), \mathcal{I}_b^{k}(t)\}$ whose components are the last action corresponding to request $k$, transmission rate history corresponding to the user $\mathcal{U}_b[k]$ of request $k$, and the information of request $k$. 

The input that the decoder LSTM receives at each step is $\mathcal{O}_b^{dc}(t) = \{H^b[\mathcal{U}_b(t), c](t), \cup_{k=1}^{n_r} \mathcal{I}_b^{k}(t)\}$. The two components are the transmission rate history corresponding to all users $\mathcal{U}_b$ in channel $c$, and the information of all requests being processed. 

Therefore as in (\ref{eq:att_ptr}), the output $e_k$ and $d_c$ of the encoder and decoder are fed into the attention mechanism to produce the conditional probability $\mathcal{P}$, and thus we obtain the output action $\mathcal{O}_b$ via (\ref{eq:p_to_act}). 

Apart from the pointer network actor $f^\theta(\mathcal{O}_b)$, we consider two other decision modes to explore different actions and train MACC policies, including the random mode and the max-rate mode. The random mode randomly assigns $n_c$ out of $N$ channels to $n_r$ requests without considering the observation, and the max-rate mode generates $\mathcal{O}_b$ via transmission rate history matrix $H^b[\mathcal{U}_b, \mathcal{C}]$ instead of $\mathcal{P}$:
\begin{equation} \label{eq:H_to_act}
\begin{aligned} 
M_1 & = [\operatorname*{max}(H^b[\mathcal{U}_b, 1]), \ldots, \operatorname*{max}(H^b[\mathcal{U}_b, N])] , \\
M_2 & = \operatorname*{argmax}_{M_1' \subset M_1, |M_1'|=n_c} {\sum_{m \in M_1'}{m}} , \\
\mathcal{A}_b[k, c] & = \begin{cases}
1 \hspace{.3cm} \text{if $c \in M_2$ and $k = \operatorname*{argmax}H^b[\mathcal{U}_b, c]$,}\\
0 \hspace{.3cm} \text{otherwise.}
\end{cases}\\
\end{aligned} 
\end{equation}

The agent follows an $\epsilon$-$\epsilon_m$-greedy policy as shown in Algorithm \ref{alg:epsilon} below to update the neural network parameters. Specifically, it initially starts with an exploration probability $\epsilon=\epsilon_0=1$ and chooses the max-rate mode with probability $\epsilon_m$, otherwise chooses the random mode. This probability decreases linearly to $\epsilon = \epsilon_1 \ll 1$ to follow mostly the actor policy $f^\theta(\mathcal{O}_b)$. When the training is over, $\epsilon$ is fixed to $\epsilon_1$ in the testing duration $T_{test}$. 

\begin{algorithm}
\small
\caption{$\epsilon$-$\epsilon_m$-greedy exploration method}
	\label{alg:epsilon}
	\begin{algorithmic}
		\State {$\epsilon=\epsilon_{0}$}
		\For{t in range($T_{train}$)}
		    \For{b in range($N_B$)}
    			\If {random$(0,1) \ge \epsilon$}
    				\State{Agent $b$ executes $f^\theta(\mathcal{O}_b)$ in (\ref{eq:p_to_act})}
    			\Else
    				\If {random$(0,1) \le \epsilon_m$}
    					\State{Agent $b$ executes max-rate mode in (\ref{eq:H_to_act})}
    				\Else
    					\State{Agent $b$ executes random mode}
    				\EndIf
			    \EndIf
			\EndFor
			\State {$\epsilon=\epsilon-(\epsilon_{0}-\epsilon_{1})/T_{train}$}
		\EndFor
	\end{algorithmic}
\end{algorithm}

\subsection{Numerical Results}\label{sec: exp}
\begin{figure}
	\centering
	\includegraphics[width=0.8\linewidth]{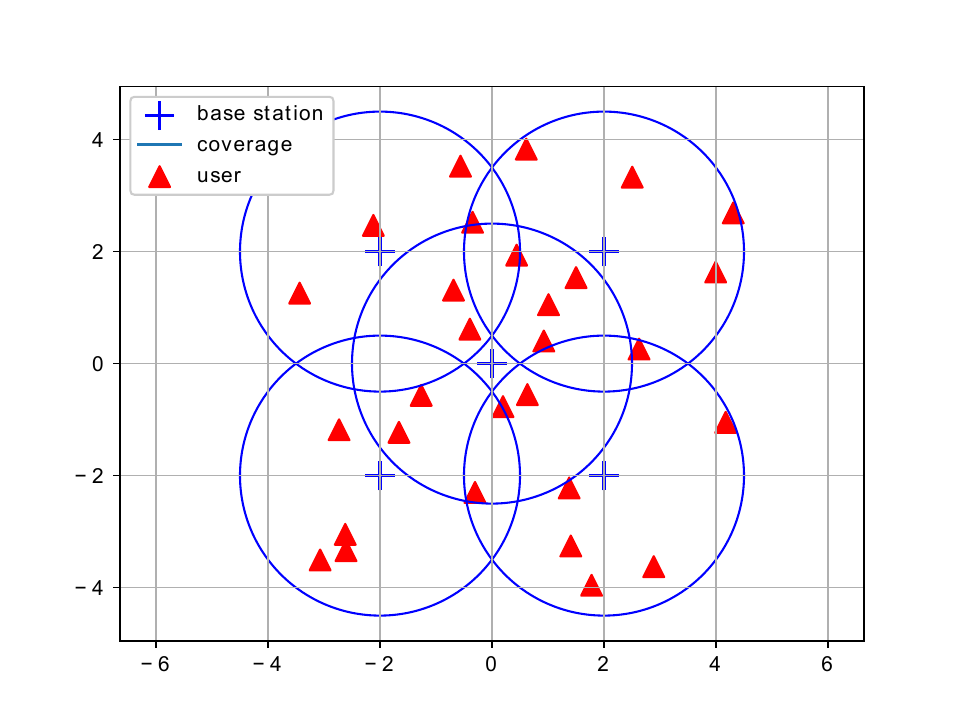}
	\caption{Coverage map of service area with 5 base stations and 30 users. }
	\label{fig:map}
\end{figure}

As shown in Fig. \ref{fig:map}, we in the experiments consider a service area with $N_B=5$ base stations and $N_u=30$ users, and each cell radius is 2.5km. There are $N=16$ channels available, and each base station picks at most $N_c=8$ channels for transmission. The ratio between transmission power and the noise in each channel is $P_B / \sigma_k^2 = 6.3$. Each base station allows the processing of $N_r=4$ requests, and keeps at most $N_q=2$ requests in the queue. When a request from one user is terminated or completed, the user will not be able to send another request within $T_r=2$ time slots. 

The channel varies with maximum Doppler frequency $f_d=1$ and dynamic slot duration $T=0.02s$, and the location of users follows a random walk pattern. Each user may transfer between the coverage of different base stations, while staying within the coverage area of at least one of the base stations. All other parameters are listed in Table \ref{tab:para}. 

\begin{table}[h!]
	\scriptsize
	\begin{center}
		\caption{User Agent Training Parameters}
		\label{tab:para}
		\begin{tabular}{ | l || l |}
			\hline
			Number of Base Stations $N_B$ & 5 \\ \hline
			Height of Base Station $h_B$ & {50 m} \\ \hline
			Range of Base Station Coverage & 2.5 {km} \\ \hline
			Number of MVNOs $N_M$ & 1 \\ \hline
			Number of Users $N_u$ & 30 \\ \hline
			Number of Channels $N$ & 16 \\ \hline
			Number of Channels for Transmission $N_c$ & 8 \\ \hline
			Maximum Doppler Frequency $f_d$ & 1 {Hz} \\ \hline
			Dynamic Slot Duration $T$ & 0.02 {ms} \\ \hline
			Transmission Power to Noise Ratio $P_B / \sigma_k^2$ & 6.3 {dBm}   \\ \hline
			Path Loss Exponent $\alpha$ & -2 \\ \hline
			Maximum Number of Requests Processing $N_r$ & 4 \\ \hline
			Maximum Number of Requests in Queue $N_q$ & 2 \\ \hline
			Minimum Inter-arrival Time of Requests $T_r$ & 2 \\ \hline
			Random Range of Initial Payload $p$ & (1, 2) \\ \hline
			Random Range of Minimum Rate $m$ & (0.8, 1) \\ \hline
			Random Range of Lifetime $l$ & $p/m+$ (2, 4) \\ \hline
			Discount Factor $\gamma$ & 0.9 \\ \hline
			Adam Optimizer Learning Rate & $10^{-6}$ \\ \hline
			Critic FNN Hidden Layer and Nodes & 1, 70 \\ \hline
			Encoder LSTM Hidden State Size $h_e$ & 70 \\ \hline
			Decoder LSTM Hidden State Size $h_d$ & 70 \\ \hline
			Training Time $T_{train}$ & 50000 \\ \hline
			Testing Time $T_{test}$ & {150000} \\ \hline
			Training Period $T_t$ & 10 \\ \hline
			Mini-batch & 10 \\ \hline
			Initial Exploration Probability $\epsilon_0$ & 1 \\ \hline
			Final Exploration Probability $\epsilon_1$ & 0.005 \\ \hline
			Max-Rate Probability $\epsilon_m$ & 0.1 \\ \hline
		\end{tabular}
	\end{center}
\end{table}

\begin{figure}
	\centering
	\includegraphics[width=1.\linewidth]{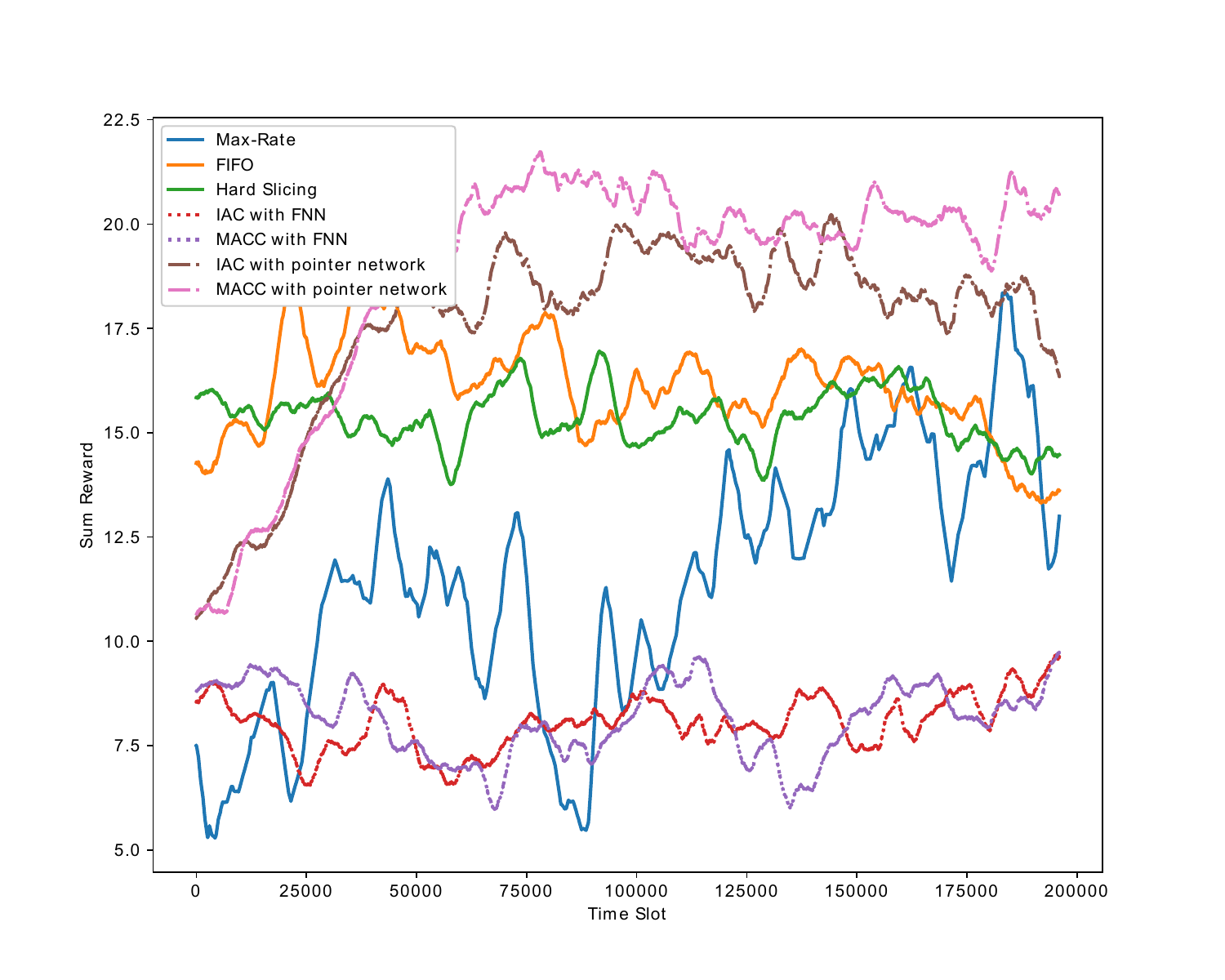}
	\caption{{Performance comparison in terms of sum reward for the proposed MACC agent with pointer network based actors against different algorithms.} }
	\label{fig:user_curve}
\end{figure}

{In Fig. \ref{fig:user_curve}, we compare the moving average of the sum reward achieved by the network slicing agents utilizing the proposed MACC with pointer networks against other algorithms. Similarly as in \cite{wang2022multi}, we introduce three statistical algorithms including max-rate which always executes the max-rate mode, FIFO that selects channels with high channel rate history but always assigns enough channel resources to the requests that has arrived earlier, and hard slicing that assigns channels with high channel rate history and the number of channels assigned to each request is proportional to the corresponding minimum transmission rate $m_k$. We also compare four deep RL agents, each of which deploys IAC or MACC frameworks using either FNN and pointer networks as actor. While the proposed MACC with pointer network agent starts with a lower performance at the beginning of the training phase due to random exploration, it outperforms all other algorithms during the test phase. In the initial training phase of $t<50000$, the proposed agent starts with random parameters and employs the $\epsilon$-$\epsilon_m$-greedy policy, which starts with a large probability $\epsilon (1-\epsilon_m)$ to explore with random actions, and consequently the performance starts low but gradually improves. We observe that in the test phase from $t=50000$ to $t=200000$, the sum reward eventually converges to around 20 bits/symbol, while slightly oscillating due to the challenging dynamic environment with random channel states, varying user locations, and randomly arriving requests. We further note that the proposed network slicing agents complete over $95\%$ of the requests, and hence attain a very high completion ratio as well. }
% In Fig. \ref{fig:user_curve}, we plot the moving average of the sum reward achieved by the network slicing agents utilizing the proposed MACC with pointer networks. During the training phase of $t<50000$, these deep RL agents start with random parameters and employ the $\epsilon$-$\epsilon_m$-greedy policy, which starts with a large probability $\epsilon (1-\epsilon_m)$ to explore with random actions, and consequently the performance starts low but gradually improves. We observe that in the test phase from $t=50000$ to {$t=200000$}, the sum reward eventually converges to around 20 bits/symbol, while slightly oscillating due to the challenging dynamic environment with random channel states, varying user locations, and randomly arriving requests.  We further note that the network slicing agents complete around $95\%$ of the requests, and hence attain a very high completion ratio as well. 

\section{Deep RL Based Jammer}\label{sec: jam}
{We have introduced the deep RL based network slicing agents in the previous section. Since deep RL is vulnerable to adversarial attacks that perturb the observations, the proposed network slicing agents can be open to attack by intelligent jammers, and it is critical to determine their sensitivity and robustness under such jamming attacks.} In this section, we introduce an actor-critic deep RL agent that performs jamming attacks on the aforementioned victim network slicing agents/users (introduced in Section \ref{sec: ac}) and aims at minimizing the victims' transmission rate. We assume the jammer has the geometric map of all BSs, but it does not have any information on the channel states, users' locations, requests, victim reward or the victim policy. This deep RL jammer may jam multiple channels to reduce the transmission rate and potentially may lead to request failures, or observe the environment and record the interference power in each channel to speculate the victims' actions. We demonstrate a jammer that can significantly degrade the aforementioned victim users' performance even though it lacks critical information on the victims.

\subsection{System Model}\label{subsec: jamsys}
In the considered channel model, the fading coefficient of the link from the jammer to the user equipment (UE) $u$ in a certain channel $c$ is denoted by $h_{c}^{J,u}$, and the fading coefficient of the link from the base station $b$ to the jammer in a certain channel $c$ is denoted by $h_{c}^{b,J}$. We consider $h_{c}^{b,u}$, $h_{c}^{J,u}$ and $h_{c}^{b,J}$ to be independent and identically distributed (i.i.d.) and vary over time according to the Jakes fading model \cite{liang2017spectrum}. Once the jammer is initialized at horizontal location $\{x_J, y_J\}$ with height $h_J$, it can choose in any given time slot one of the two operational phases: jamming phase and listening phase.

\subsubsection{Jamming phase}
In this phase, the jammer jams $n_J \le N_J$ channels simultaneously with jamming power $P_J$ in each channel without receiving any feedback. With the additional interference from the jammer, we can express the transmission rate $\mathsf{r}_{c,J}^{b,u}$ from base station $b$ to UE $u$ in channel $c$ as
\begin{equation} \label{eq:individual_rate_jam}
\mathsf{r}_{c,J}^{b,u} = \log_2 \left( 1+{\frac{P_B L^{b,u} |h_{c}^{b,u}|^2}{\sum_{b' \neq b}{\mathcal{N}^{b, b'}_c} + \mathcal{N}^{b, J}_c + \sigma^2}} \right) ,
\end{equation}
\begin{equation} \label{eq:jam_intf}
\mathcal{N}^{b, J}_c = \mathbf{1}_c^{b,J} P_J L^{J,u} |h_{c}^{J,u}|^2 ,
\end{equation}
where $P_J$ is the jamming power, $\mathcal{N}^{b, J}_c$ is the jamming interference in channel $c$, $\mathbf{1}_c^{b,J}$ is the indicator function for both base station $b$ and jammer choosing channel $c$, and $L^{J,u}$ is the path loss:
\begin{equation} \label{eq:pathloss_Ju}
L^{J,u} = \left(h_J^2 + (x_J - x_u)^2 + (y_J - y_u)^2\right)^{\alpha/2}.
\end{equation}
%In (\ref{eq:individual_rate_jam}), $P_B$ is the transmission power of the base stations, $h_{c}^{b,u}$ is the fading coefficient of the link between base station $b$ and UE $u$ in channel $c$, $\mathcal{N}^{b, b'}_c$ is the interference from base station $b'$ in channel $c$ (if  base station $b'$ is also transmitting in channel $c$ based on the network slicing decisions), $\sigma^2$ is the variance of the additive white Gaussian noise, and $L^{b,u}$ is the path loss in the link between base station $b$ and UE $u$
% \begin{equation} \label{eq:pathloss}
% \begin{aligned} 
% L^{b,u} & = \left(h_B^2 + (x_b - x_u)^2 + (y_b - y_u)^2\right)^{\alpha/2}
% \end{aligned}
% \end{equation}
% where $h_B$ is the height of each base station, $\alpha$ is the path loss exponent, and $\{x_b, y_b\}$ and $\{x_u, y_u\}$ are the 2-D coordinates of base station $b$ and UE $u$, respectively. 

By degrading the transmission rate $\mathsf{r}_{c,J}^{b,u}$ with the jamming interference in channel $c$, the jamming attack may lead to a number of request failures. The jammer may further amplify its impact by intelligently choosing a preferable subset of channels to jam. The listening phase is introduced to learn such information from the environment. 

\subsubsection{Listening phase}
In this phase, the jammer does not jam any channel, but only listens the (interference) power in each channel $c$ among $N$ channels:
\begin{equation} \label{eq:listen_intf}
\mathcal{N}^\text{listen}_c = {\sum_{b}{\mathbf{1}_c^{b} P_B L^{b, J} |h_{c}^{b, J}|^2} + \sigma^2} ,
\end{equation}
where $\mathbf{1}_c^{b}$ is the indicator function which has a value of $1$ if there is a transmission at base station $b$ to any UE in channel $c$, and $L^{b, J}$ is the path loss:
\begin{equation} \label{eq:pathloss_bJ}
L^{b, J} = \left((h_B - h_J)^2 + (x_b - x_J)^2 + (y_b - y_J)^2\right)^{\alpha/2}.
\end{equation}

Due to the jammer's lack of prior information, we consider an approximation 
%where for each user-base-station pair, $\{x_u, y_u\} = \{x_B, y_B\}$, $h_B=0$, and $|h_{c}^{J,u}| = |h_{c}^{b, J}|$. 
and assume that the listened power $\mathcal{N}^\text{listen}_c$ from all base stations transmitting in channel $c$ is a rough estimate of the sum of jamming interferences $\mathcal{N}^{b, J}_{c,\text{est}}$ if jammer were in the jamming phase and  chose channel $c$ to inject interference, i.e.,
\begin{equation} \label{eq:est_intf}
\mathcal{N}^\text{listen}_c \approx {\sum_{b}{\mathcal{N}^{b, J}_{c,\text{est}}} + \sigma^2} .
\end{equation}
Therefore, with this assumption, the jammer anticipates that the higher $\mathcal{N}^\text{listen}_c$ being observed/listened, the more likely that jamming in channel $c$ degrades the victim users' performance. Given this, we  introduce how we optimize the subset of channels to attack during the jamming phase in Section \ref{subsec: jamch}.

Another benefit of the listening phase is that no jamming power is consumed in this phase, and consequently average power consumption is reduced. In the remainder of this paper, we assume that the jammer only switches from the listening phase to the jamming phase by the end of each period with $T_J \in \mathbb{R}^+$ time slots, and thus it has an average power consumption of $n_J P_J / T_J$.

\subsection{Jamming Location Optimization}\label{subsec: jamloc}
The jammer aims at minimizing the performance of victim users, but it does not have any information on channel fading, UE locations or rewards provided to different requests. Therefore, the jamming location is optimized by minimizing the expected sum transmission rate for given UE $u$ integrated over the service area when the channels for transmission coincide with the channels being jammed. More specifically, we have the following optimization:
\begin{equation} \label{eq:jam_loc}
\{x_J^*, y_J^*\}=\operatorname*{argmin}_{\{x_J, y_J\}}{
\mathbb{E}_{h} \left( \sum_b \sum_c \iint\limits_{D^b_h}{\mathsf{r}_{c,J}^{b,u} d x_u d y_u} \right)
},
\end{equation}
where the expectation with respect to the set of fading coefficients $\{h\}$ considers $\forall b, u, c: h_{c}^{b,u}, h_{c}^{J,u}, h_{c}^{b,J} \overset{\text{i.i.d.}}{\sim} \mathcal{CN} (0,1)$, $D^b_h$ is the subset of coverage area with maximal transmission rate $\mathsf{r}_{c,J}^{b,u}$ from base station $b$ given $\{h_{c}^{b,u}, h_{c}^{J,u}, h_{c}^{b,J}\}$, and $\forall b,b': \mathbf{1}_c^{b,b'}=0, \mathbf{1}_c^{b,J}=\mathbf{1}_c^{b}=1$. Additionally, we note that $P_B$, $h_B$, $|\alpha|$, and $\sigma$ with arbitrary positive values will not affect the optimized jamming location.

\subsection{Jamming Channel Optimization}\label{subsec: jamch}
After the jammer is initialized in the true environment and has observed/listened the interference power $\mathcal{N}^\text{listen}_c(t-1)$ within the listening phase at time $t-1$, it will decide the subset of channels $\mathcal{C}_J(t) \subset \mathcal{C}$ to jam during jamming phase at time $t$, where $\left| \mathcal{C}_J(t) \right| = n_J(t)$ and $\mathcal{C}$ is the set of all channels $\{1, 2, \ldots, N\}$. According to (\ref{eq:est_intf}), channels with higher $\mathcal{N}^\text{listen}_c$ are more likely to be better choices, but this information is not available in the jamming phase at time $t$ (since jamming is performed rather than listening). Therefore, $\mathcal{C}_J(t)$ can only be evaluated via $\mathcal{N}^\text{listen}_c(t-1)$ and $\mathcal{N}^\text{listen}_c(t+1)$. Note that $\mathcal{N}^\text{listen}_c(t+1)$ is not available at time $t$ and this challenge will be addressed via deep RL in the next subsection (i.e., by essentially introducing a reward  to train the neural network at time $t+1$ and having that reward depend on $\mathcal{N}^\text{listen}_c(t-1)$ and $\mathcal{N}^\text{listen}_c(t+1)$.) Hence, with the deep RL approach, the action will depend only on observation before time $t$. 

In the absence of information on the requests, we assume a model in which each request arrives and is completed independently. Thus, the state of current time slot $t$ can be estimated as a linear interpolation (or a weighted average) of $\mathcal{N}^\text{listen}_c(t-1)$ and $\mathcal{N}^\text{listen}_c(t+1)$. Therefore, the optimized subset of channels to jam can be determined from
\begin{equation} \label{eq:jam_ch}
\mathcal{C}_J^*(t) = \operatorname*{argmax}_{\mathcal{C}_J} { \sum_{c \in \mathcal{C}_J}{\hat{\mathcal{N}}^{\beta}_c(t)}},
\end{equation}
where
\begin{equation} \label{eq:jam_diff}
\hat{\mathcal{N}}^{\beta}_c(t) = \frac{1}{\beta(t)+1} \left( \beta(t) \mathcal{N}^\text{listen}_c(t-1) + \mathcal{N}^\text{listen}_c(t+1) \right) ,
\end{equation}
and $\beta(t)$ describes the impact of the jammer on the victims. Typically, a request takes multiple time slots to get completed. When it is jammed, there are two possibilities. On the one hand, it may fail to meet the minimum transmission rate limit and get terminated immediately. In such a case, the next request in the queue is processed, and the network slicing agent rearranges and distributes channels into a new set of slices to be allocated to different requests from different users, and thus the listened interference $\mathcal{N}^\text{listen}_c(t+1)$ may change dramatically. In this case, we are likely to have $\beta(t) > 1$. On the other hand, if the request under attack has a lower transmission rate but the minimum transmission rate limit and the lifetime constraint are satisfied, the transmission will last longer, and the interference $\mathcal{N}^\text{listen}_c(t+1)$ is less likely to vary from that at time $t$. In this case, we are likely to have $\beta(t) < 1$. Therefore, the value of $\beta(t)$ should be determined via experience:
\begin{equation} \label{eq:jam_beta}
\beta(t) = \operatorname*{max} \left(\frac{2 |\mathcal{T}'| \sum_c{ \sum_{t'' \in \mathcal{T}''}{d^\text{listen}_c(t'') }}}{
|\mathcal{T}''| \sum_c{\sum_{t' \in \mathcal{T}'}{d^\text{listen}_c(t') }}} - 1 , 0\right) ,
\end{equation}
where 
\begin{equation}\label{eq:jam_beta_diff}
d^\text{listen}_c(t) = \left| \mathcal{N}^\text{listen}_c(t+1) - \mathcal{N}^\text{listen}_c(t-1)\right|,
\end{equation}
and $\mathcal{T}''$ is a set of time points in the jamming phase where each $t'' \in \mathcal{T}''$ is close to time $t$, and $\mathcal{T}'$ is a set of time points where each $t' \in \mathcal{T}'$ are in successive listening phases without jamming attack. If $T_J>3$, $d^\text{listen}_c(\mathcal{T}')$ can be the set of successive listening phases in every period during training. Otherwise if $T_J \le 3$, $d^\text{listen}_c(\mathcal{T}')$ has to be collected before jamming starts. 

Again, it is important to note that when the jammer agent makes decisions at time $t$, $\mathcal{N}^\text{listen}_c(t+1)$ is not available. To address this, we propose a deep RL agent that uses the actor-critic algorithm to learn the policy.

\subsection{Actor-Critic Jammer Agent}\label{subsec: jamDRL}
Our proposed jammer agent utilizes an actor-critic deep RL algorithm to learn the policy that optimizes the output $\mathcal{C}_J(t)$ to minimize the victims' expected sum rate. The jammer works with a period $T_J$, and only switches from the listening phase to the jamming phase at the end of each period and uses the policy to make the decision $\mathcal{C}_J(t)$. %The actor-critic algorithm \cite{peters2008natural} includes two neural networks, namely the actor and critic. The two networks have separate neurons and utilize separate back-propagation, and they may have separate hyper parameters. 
We next introduce the observation, action, reward, and the actor-critic update of this agent. 

\subsubsection{Observation}
At each time slot, the jammer records its instantaneous observation as a vector $O_J \in \mathbb{R}^{N}$. In a listening phase, $O_J^L=\{ \mathcal{N}^\text{listen}_1, \mathcal{N}^\text{listen}_2, \ldots, \mathcal{N}^\text{listen}_N \}$. Otherwise, in a jamming phase, $O_J^J=\{ \mathbf{1}(1 \in \mathcal{C}_J), \mathbf{1}(2 \in \mathcal{C}_J) \ldots, \mathbf{1}(N \in \mathcal{C}_J) \}$ where $\mathbf{1}$ is the indicator function. In the beginning at time slot $t$ in the jamming phase, the full observation $\mathcal{O}_J(t) = \{ O_J^J(t-T_J), O_J^L(t-T_J+1), \ldots, O_J^L(t-1) \}$ is fed as the input state to the actor-critic agent.

\subsubsection{Action}
At the beginning of time slot $t$ in a jamming phase, given the input state $\mathcal{O}_J(t)$, the actor neural network outputs a vector of probabilities $\mathcal{P}_J(t) \in [0, 1]^N$. From the probability vector, the decision $\mathcal{C}_J(t)$ is derived which is the subset of channels to jam, and it is described as the action $\mathcal{A}_J(t) \in \{0, 1\}^N$:
\begin{equation} \label{eq:jam_act_C}
\mathcal{C}_J(t) = \operatorname*{argmax}_{\mathcal{C}_J} { \sum_{c \in \mathcal{C}_J}{\mathcal{P}_J(t)}},
\end{equation}
\begin{equation} \label{eq:jam_act_A}
\mathcal{A}_J(t)=\{ \mathbf{1}(1 \in \mathcal{C}_J(t)), \mathbf{1}(2 \in \mathcal{C}_J(t)) \ldots, \mathbf{1}(N \in \mathcal{C}_J(t)) \}.
\end{equation}

\subsubsection{Reward}
Following the jamming phase at time $t$, the reward is received to train the critic after the next listening phase at time $t+1$. This reward aims at encouraging the policy to produce an action that imitates $\hat{\mathcal{N}}^{\beta}_c(t)$, the linear interpolation of listened interference as in (\ref{eq:jam_diff}). Therefore, we set the reward as the negative of the mean squared error:
\begin{equation} \label{eq:jam_reward}
R_J(t) = -\sum_{c}{\left(\mathcal{A}_J(t)[c] - \frac{\hat{\mathcal{N}}^{\beta}_c(t)}{\operatorname*{max} \left( \hat{\mathcal{N}}^{\beta}_c(t) \right) } \right)^2}.
\end{equation}

\subsubsection{Actor-Critic Update}
At the beginning of a jamming phase at time $t$, the actor with parameter $\theta_J$ and policy $f^{\theta_J}(\mathcal{O}_J)$ maps the input observation $\mathcal{O}_J$ to the output probability $\mathcal{P}_J$, which is similar to a Q-value generator. The critic with parameter $\phi_J$ and policy $g^{\phi_J}(\mathcal{O}_J)$ maps $\mathcal{O}_J$ to a single temporal difference (TD) error:
\begin{equation} \label{eq:acdelta_jam}
\delta_J(t) = {R_J}(t) + \gamma_J g^{\phi_J}(\mathcal{O}_J(t)) - g^{\phi_J}(\mathcal{O}_J(t-T_J)),
\end{equation}
where $\gamma_J \in (0, 1)$ is the discount factor. For each training sample, the critic is updated towards achieving the optimized parameter $\phi_J^*$ to minimize the least square TD:
\begin{equation} \label{eq:lstd_jam}
\phi_J^*=\operatorname*{argmin}_{\phi_J}{(\delta_J^{g_{\phi_J}})^2}.
\end{equation}
The actor is updated towards the optimized parameter $\theta_J^*$ to minimize the policy gradient:
\begin{equation} \label{eq:pg_jam}
\theta_J^*=\operatorname*{argmax}_{\theta_J}{\nabla_{\theta_J} \sum_{c \in \mathcal{C}_J} \log f^{\theta_J} (\mathcal{O}_J) \delta_J^{g_{\phi_J}}} .
\end{equation}
Both networks are updated alternately to attain the optimal actor-critic policy. Note that during each parameter update with the bootstrap method, the mini-batch of training samples (i.e., action-reward pairs) should be randomly drawn from a longer history record for faster convergence.  

\subsection{Numerical Results}\label{subsec: jamexp}
\begin{figure}
	\centering
	\includegraphics[width=1\linewidth]{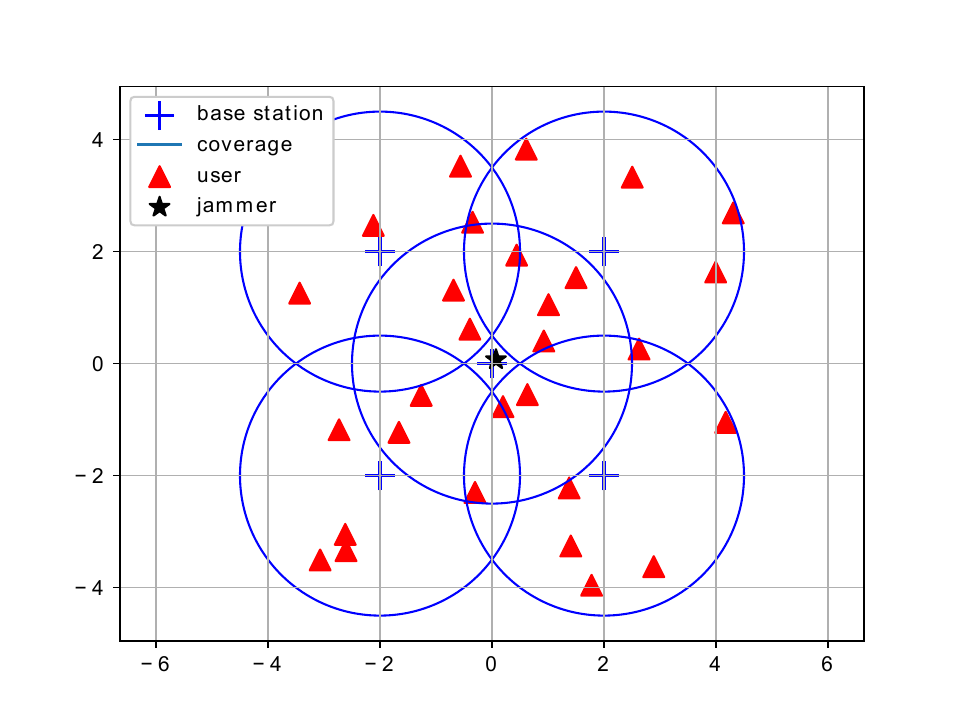}
	\caption{Coverage map of service area with 5 base stations, 30 users, and the jammer. }
	\label{fig:map_jam}
\end{figure}

As shown in Fig. \ref{fig:map_jam}, we in the experiments consider a service area with $N_B=5$ base stations, $N_u=30$ users, and a jammer. The theoretically optimized location of the jammer $\{x_J^*, y_J^*\}$ is determined according to (\ref{eq:jam_loc}) and it lies at the center $\{0, 0\}$, while the actual location is slightly moved away from the base station tower. There are $N=16$ channels available, and the jammer picks at most $N_J=8$ channels for jamming. The jamming power in each channel equals the transmission power in each channel: $P_J = P_B$. The jammer has phase switching period of $T_J=2$ time slots.

In the experiments, the network slicing agents are initialized as the well-trained MACC agents with pointer network actors as detailed in Section \ref{sec: ac}, and time is initiated at $t=0$. The jammer's actor and critic policies are implemented as two feed-forward neural networks (FNNs), and both have one hidden layer with 16 hidden nodes. The jammer is initialized at $t=100$ and begins jamming and performs online updating. During the training phase $100 \le t < 10000$, the jammer follows an $\epsilon$-greedy policy to update the neural network parameters $\theta_J$ and $\phi_J$ with learning rate $10^{-5}$. It starts by fully exploring random actions, and the probability to choose random actions linearly decreases to $0.01$, thus eventually leading the agent to mostly follow the actor policy $f^{\theta_J}(\mathcal{O}_J)$. This probability is fixed to $0.01$ in the testing phase from $t=10000$ to $t=20000$.  

\begin{figure}
	\centering
	\includegraphics[width=1\linewidth]{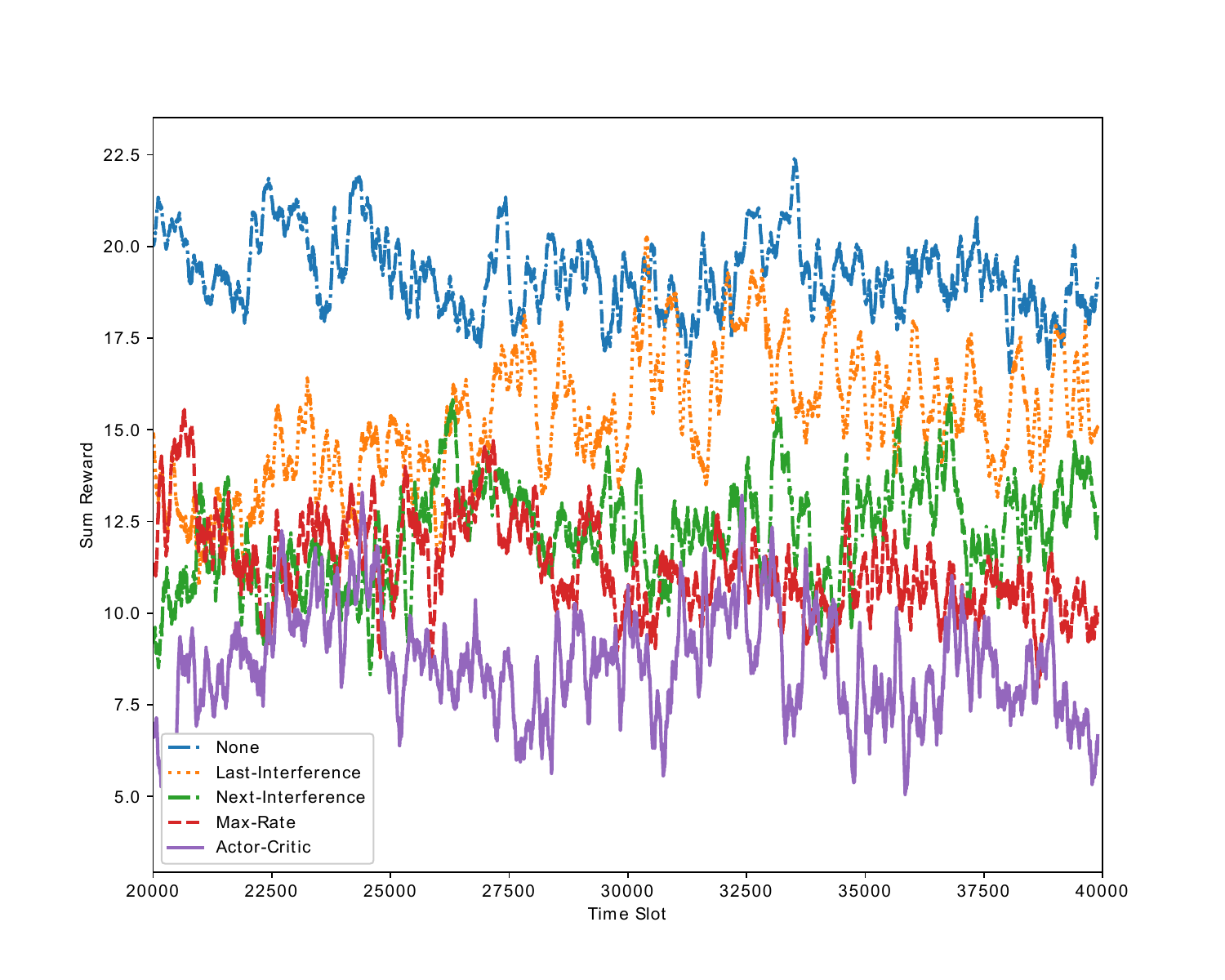}
	\caption{{Comparison of victims' sum reward in the testing phase achieved in the absence of jamming attack, and also achieved under attacks by the last-interference jammer, next-interference jammer, and the proposed actor-critic jammer. The jamming power equals the transmission power.}}
	\label{fig:jam_curve}
\end{figure}

\begin{figure}
	\centering
	\includegraphics[width=1\linewidth]{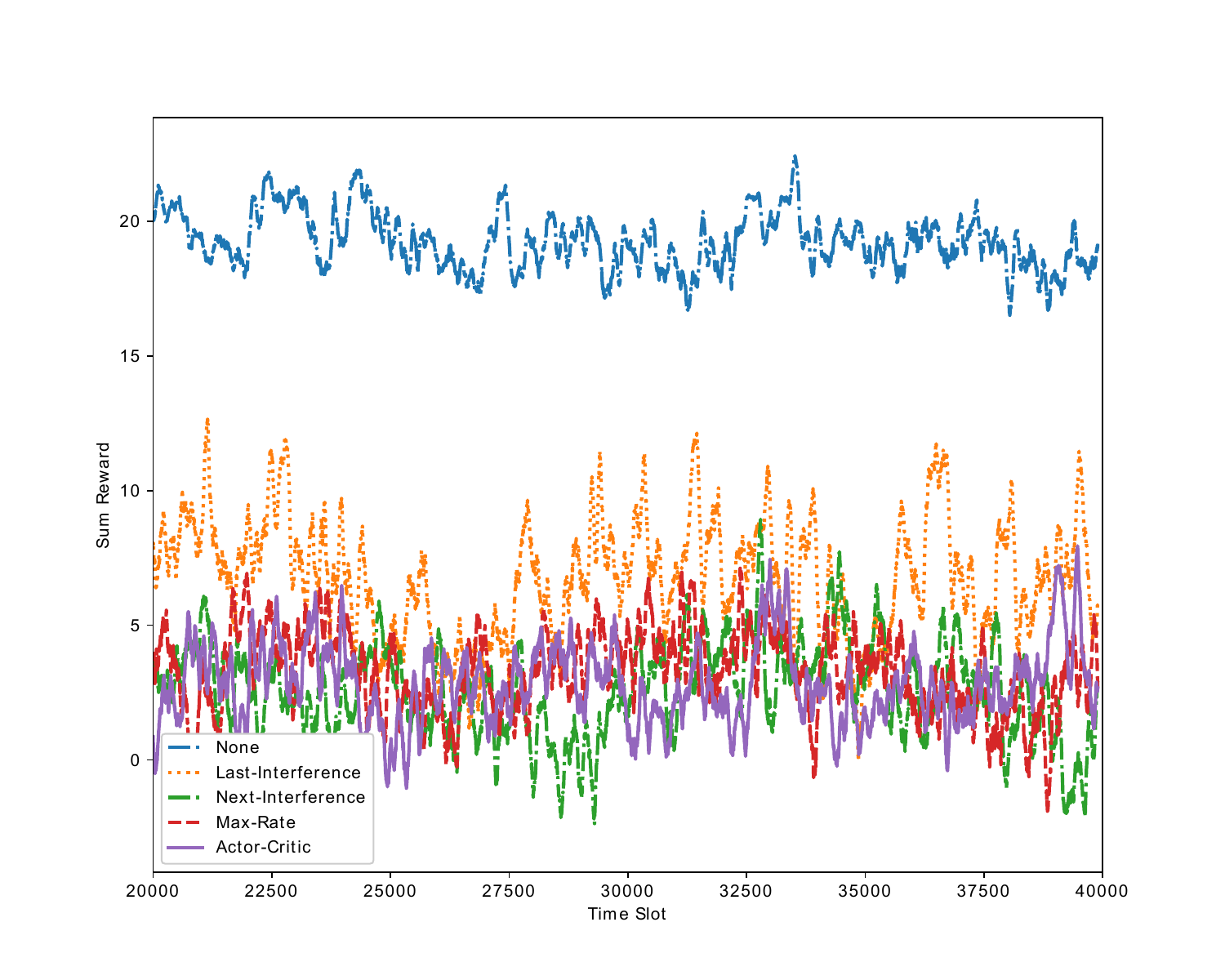}
	\caption{{Comparison of victims' sum reward in the testing phase achieved in the absence of jamming attack, and also achieved under attacks by the last-interference jammer, next-interference jammer, and the proposed actor-critic jammer. The jamming power is 60dB higher than the transmission power.}}
	\label{fig:jam_curve_60dB}
\end{figure}

In Fig. \ref{fig:jam_curve}, we compare the performance of the proposed actor-critic jammer that approximates $\hat{\mathcal{N}}^{\beta}_c(t)$ with {four} other scenarios in terms of victim sum reward in the testing phase. {The figure shows the testing phase after the victim agent adapts to the jamming attack, and as a result its action policy becomes stable.  The slight fluctuations in the curves are due to randomly arriving requests and varying channel states. We also note that the vertical axis is the victim's reward, and hence  the jammer with the better performance will lead to lower victim reward. } The first case is the setting with no jammer and hence the performance is that of the original network slicing agent in terms of the sum reward in the absence of jamming attacks. The second scenario is with a last-interference jammer agent that is positioned at the same location (i.e., the origin $\{0,0\}$) with the same power budget. However, this jammer agent does not utilize any machine learning algorithms, and chooses the subset of channels with the highest observed/listened interference power levels in the last listening phase: 
\begin{equation} \label{eq:jam_ch_maxintf}
\mathcal{C}_J^{\text{MaxIntf}}(t) = \operatorname*{argmax}_{\mathcal{C}_J} { \sum_{c \in \mathcal{C}_J}{
\mathcal{N}^\text{listen}_c(t-1) 
}}.
\end{equation}
Consequentially, the last-interference jammer which focuses on the last time slot is equivalent to the proposed jammer when $\beta \rightarrow \infty$. The third scenario is with the next-interference jammer agent, which is also an actor-critic jammer agent but whose reward has $\beta = 0$, so it concentrates on the next time slot. {Furthermore, we provide performance comparisons with a max-rate jammer that is assumed to know the channel environment between the base stations and users perfectly (which implies a rather strong jammer), and thus it obtains every potential channel rate regardless of the interference from other users, and picks the maximum via
\begin{equation} \label{eq:jam_maxrate}
r^{max}_c = \operatorname*{max}_{b, u} { \log_2 \left( 1+{\frac{P_B L^{b,u} |h_{c}^{b,u}|^2}{\sigma^2}} \right)}.
\end{equation}
Given $N$ maximum potential rates in $N$ different channels, the jammer randomly picks channel $c$ to jam with probability $r^{max}_c / \sum_{c'}^N{r^{max}_{c'}}$. To better evaluate the performance among different jammers, we assume that all jammers have the same power budget. We observe that all other jammers} are less efficient in suppressing the victim sum reward compared to our proposed actor-critic jammer, which aims at estimating  $\hat{\mathcal{N}}^{\beta}_c(t)$ and therefore has better performance. Specifically, we observe in Fig. \ref{fig:jam_curve} that the proposed jammer results in the smallest sum reward values for the network slicing agents and has the most significant adversarial impact. {In order to give a comprehensive comparison, we in Fig. \ref{fig:jam_curve_60dB} also illustrate the performance of the same set of jammers where the jamming power is 60dB higher than the transmission power, i.e. $P_J = {10}^{6} P_B$. We notice that since the victim user receives negative reward when the request fails, some curves have negative values at times in this harsh jamming interference environment.}

% \begin{figure}
% 	\centering
% 	\includegraphics[width=1\linewidth]{jammer_plot}
% 	\caption{Comparison of victims' sum reward in the testing phase achieved in the absence of jamming attack, and also achieved under attacks by the last-interference jammer, next-interference jammer, and the proposed actor-critic jammer. }
% 	\label{fig:jam_curve}
% \end{figure}

% Furthermore, we show the numerical result of the same set of experiments in Table \ref{tab:perform_jam}, where we observe that the victim under the actor-critic jamming attack completes only 73.68\% of the requests, which is much less than the other cases. Additionally, we notice that the base station at coordinates $(0, 0)$ in Fig. \ref{fig:map_jam}, which is the closest one to the jammer's location, only completes 67.25\% of the requests under the proposed jamming attack. In comparison, this base station under the other two types of attack completes about 80\% of the requests. This indicates that the proposed agent is more likely to learn from the received interference. 
{While it is hard to distinguish some curves with similar performance, we provide the numerical results of average reward and request completion ratio in both experiments in Table \ref{tab:perform_jam}. We observe that the victim under the actor-critic jamming attack completes much less requests and attains the smallest average reward values in both scenarios, and thus the actor-critic jammer outperforms all other jammers including the max-rate jammer that knows the channel status. Additionally in the first scenario where $P_J = P_B$, we notice that the base station at coordinates $(0, 0)$ in Fig. \ref{fig:map_jam}, which is the closest one to the jammer's location, only completes 67.25\% of the requests under the proposed jamming attack. In comparison, this base station under other types of attack completes about 80\% of the requests. This base station also completes less requests under our proposed attack when $P_J = {10}^{6} P_B$. These observations indicate that the proposed jammer agent is more likely to learn from the received interference. }

% \begin{table}[h!]
% 	%\small
% 	\begin{center}
% 		\caption{Performance Comparison of Different Jamming Algorithms during Test Phase}
% 		\label{tab:perform_jam}
% 		\begin{tabular}{ | l || l | l | }
% 			\hline
% 			jammer & average reward & completion ratio  \\ \hline\hline
% 			None & 19.90 & 94.78\%  \\ \hline
% 			Last-Interference & 13.94 & 82.36\%  \\ \hline
% 			Next-Interference & 14.51 & 83.30\%  \\ \hline
% 			Actor-Critic & 10.09 & 73.68\%  \\ \hline
% 		\end{tabular}
% 	\end{center}
% \end{table}
\begin{table}[h!]
	%\small
	\begin{center}
		\caption{{Performance Comparison on average reward and completion ratio of Different Jamming Algorithms during Test Phase}}
		\label{tab:perform_jam}
		\begin{tabular}{ | l || l | l | }
			\hline
			jammer & $P_J = P_B$ & $P_J = {10}^{6} P_B$  \\ \hline\hline
			None & 19.47, 93.33\% & 19.47, 93.33\%  \\ \hline
			Last-Interference & 14.82, 82.36\% & 7.23,$\,\  $ 75.30\%  \\ \hline
			Next-Interference & 11.67, 81.34\% & 3.83,$\,\  $ 68.73\% \\ \hline
            Max-Rate & 11.03, 76.07\% & 3.61,$\,\  $ 63.75\%   \\ \hline
			Actor-Critic & \textbf{9.16},$\,\  $ \textbf{73.89\%} & \textbf{2.61},$\,\  $ \textbf{60.62\%}  \\ \hline
		\end{tabular}
	\end{center}
\end{table}

\section{Nash-Equilibrium-Supervised Policy Ensemble for Jamming and Defense}\label{sec: equil}
As demonstrated in the previous section, an actor-critic based jammer with limited information and power budget %less than a base station 
can lead to significant degradation in the performance of network slicing. In this section, instead of jamming detection or jamming pattern-based defensive strategies, we  propose the Nash-equilibrium-supervised policy ensemble (NesPE) as a strategy that automatically explores the underlying available actions to achieve the optimal performance in both competitive and non-competitive contexts. We show that NesPE can be both utilized as  the victims' defensive strategy and also applied to jamming attack design.

\subsection{Nash-Equilibrium-Supervised Policy Ensemble} \label{subsec: equil}
We first describe the details of the NesPE as a scheme that enhances the performance of a player in a multi-player zero-sum stochastic game \cite{shapley1953stochastic} in which the player does not have direct observation of the opponents' action choices. The actions of the optimal mixed strategy depend on the prior observation, and this relationship is to be determined via deep RL or other machine-learning-based algorithms. To train the policy ensemble with the deep RL algorithm of the player and to determine the optimal mixed strategy profile, we consider the tuple $\mathcal{G} = \left( \mathcal{O}, \mathcal{E}, (f^{\theta_e})_{e \in \mathcal{E}}, \hat{e}, \mathcal{L}, (\mathcal{O}_l)_{l \in \mathcal{L}}, u \right)$, where
\begin{itemize}
    \item $\mathcal{O}$ is the prior observation before decision-making.
    \item $\mathcal{E}$ is a finite index set of policies, i.e., $\mathcal{E} = \{1, \ldots, E\}$.
    \item $f^{\theta_e}$ is the actor function of policy $e$ with parameter $\theta_e$, and is regarded as the player's strategy. It maps the prior observation to the player's action $\mathcal{A}_e = f^{\theta_e}(\mathcal{O}) \in \{0, 1\}^{N}$ where the chosen elements have value $1$ and others are set to $0$. 
    \item $\hat{e}$ is the index of the chosen policy to execute the action.
    \item $\mathcal{L}$ is the finite index set of subsequent observations obtained after decision-making, i.e. $\mathcal{L} = \{1, \ldots, L\}$. 
    \item $\mathcal{O}_l$ is the subsequent observation, and $l$ indirectly represents the opponents' action choice. Thus it is regarded as the opponents' strategy.
    \item $u: f^{\theta_e}(\mathcal{O}) \times \mathcal{O}_l \rightarrow \mathbb{R}$ is the utility (or payoff, reward) function of the player when it chooses policy $e$ and the interaction with the opponents results in the observation $\mathcal{O}_l$. The mapping function is unknown to both players, and its outcome is available only to the considered player after decision-making. 
\end{itemize}

According to \cite{nash1950equilibrium}, every finite strategic-form game has a mixed strategy equilibrium. In this stochastic game without prior knowledge of the utility matrix, the considered player can alternatively obtain the utility matrix by recording the experienced utility and taking the average. We consider a matrix $\mathcal{H}$ of queues with finite length (specifically,  each element of this matrix is a queue with a few recorded utilities), and the matrix has $E$ rows and $L$ columns. In each time slot, the experienced utility $u_{\hat{e},l}(t)$ is appended to the queue $\mathcal{H}(\hat{e}, l)$. At the beginning of the next time slot, the player uses the average of each queue to form a new matrix with $E \times L$ elements as the utility matrix for calculating the optimal profile at Nash equilibrium $\sigma = \{\sigma_1, \ldots, \sigma_E \}$, which is the set of probabilities to choose and execute each strategy at a time. 

However, although the probability to choose each given policy is optimized, the parameters of the policies $({\theta_e})_{e \in \mathcal{E}}$ may not be optimized, and need further training with the received utility $u_{\hat{e},l}(t)$ after each time slot $t$. Most existing works of policy ensemble use similar rewards to train different policies, thus there is no guarantee that the policy ensemble explores different strategies. In NesPE, we consider a dual problem that takes both utility reward and the correlation of policy ensemble \cite{wang2021resilient} into account. The expected correlation between policy $e_1$ and policy $e_2$ is defined as
\begin{equation} \label{eq:pol_corr}
\Bar{\rho}(e_1, e_2) = \mathbb{E}_{\mathcal{O}} \left( \sum_{n=1}^N \mathbf{1} \left( f^{\theta_{e_1}}(\mathcal{O})(n) = f^{\theta_{e_2}}(\mathcal{O})(n) = 1 \right) \right).
\end{equation}
Therefore, the correlation between the two policies indicates the similarity of decision-making patterns between them. To fully explore all possible strategies and increase the policy variety to feed the mixed strategy, lower correlations between the policies are desired. 

Thus, in order to maximize the expectation of utility reward $u_{\hat{e},l}$ and limit the correlation below a certain threshold $D$, we at each time slot $t$ consider a non-linear programming problem for policy $e$:
\begin{align}
\max_{\theta_e} \;\;\; &  \mathbb{E} \left( u_{e,l}(t) \right)  \notag \\
\text{subject to} \;\;\; & D - \sum_{e' \ne e}{\sigma_{e'}(t) \rho(e, e', t) } \ge 0,
\end{align}
where 
\begin{equation} \label{eq:corr_diff}
\rho(e, e', t) = \sum_{n=1}^N \mathbf{1} \left( f^{\theta_{e}}(\mathcal{O}(t))(n) = f^{\theta_{e'}}(\mathcal{O}(t))(n) = 1 \right).
\end{equation}
To solve the problem, we can convert it to maximizing a Lagrangian dual function, and set the reward of the deep RL algorithm as
\begin{equation} \label{eq:corr_reward}
% R(e, t) = u_{e,l}(t) + \zeta \left( D- \sum_{e' \ne e}{\sigma_{e'}(t) \rho(e, e', t) } \right).
R(e, t) = u_{\hat{e},l}(t) - \zeta \sum_{e' \ne e}{\sigma_{e'}(t) \rho(e, e', t) },
\end{equation}
where $\zeta$ is the dual variable, and $R(e, t)$ is appended to the replay buffer for the training process to update the parameters of policy $e$. We note that the instantaneous correlation $\rho(e, e', t)$ is weighted by the strategy profile $\sigma_{e'}$ of the other policy $e'$. Therefore, the more desired policies with higher $\sigma_{e}$ are less affected by the correlation with other policies, while the less desired policies are encouraged to diverge from the former. 

The proposed NesPE strategy is obtained by iteratively repeating the aforementioned process over time as shown in Algorithm \ref{alg:NesPE}.

\begin{algorithm}
\caption{Nash-Equilibrium-Supervised Policy Ensemble (NesPE)}
	\label{alg:NesPE}
	\begin{algorithmic}
	    \State{Randomly initialize the parameter $\theta_{e}$ of each policy}
	    \State{Randomly initialize the utility history $\mathcal{H}$}
	    \For{$t$ in range%($\infty$)
     }
	        \State{Acquire prior observation $\mathcal{O}(t)$}
	        \State{Obtain mixed strategy profile $\sigma$ from the average of $\mathcal{H}$}
	        \State{Randomly choose one policy $\hat{e}$ according to $\sigma$}
	        \State{Execute action $\mathcal{A}_{\hat{e}} = f^{\theta_{\hat{e}}}(\mathcal{O}(t))$}
	        \State{Classify the latter observation $\mathcal{O}_l(t)$ into category $l$}
	        \State{Append the payoff $u_{\hat{e},l}(t)$ to the queue $\mathcal{H}(\hat{e}, l)$}
	        \State{Append the information at time $t$ to the replay buffer $\mathcal{B}$}
	        \State{Randomly sample a mini-batch $\mathcal{B}'$ from $\mathcal{B}$}
	        \For{$\mathcal{G}'$, $\mathcal{A}_{\hat{e}'}$, $u_{\hat{e}',l}(t')$ in $\mathcal{B}'$}
    	        \For{$e'$ in $\mathcal{E}'$}
    	            \State{update $\theta_{e'}$ with reward $R(e', t')$ {in (\ref{eq:corr_reward})}}
    	        \EndFor
    	    \EndFor
	    \EndFor
	\end{algorithmic}
\end{algorithm}

We analyze the performance of NesPE in two contexts. On the one hand, we consider a set of opponents and an environment in which a dominating action exists, where an action $\mathcal{A}$ is called dominating if
\begin{equation} \label{eq:no_domi}
u(\mathcal{A},l) \ge u(\mathcal{A}',l), \forall \mathcal{A}' \neq \mathcal{A}, \forall l \in \mathcal{L}. 
\end{equation}
Due to random initialization, there will be one policy $e$ in NesPE that has the highest $\sigma_e$, and thus we have $|R(e, t) - u_{e,l}(t)| < |R(e_1, t) - u_{e_1,l}(t)|, \forall e_1 \neq e$. Therefore the reward of policy $e$ focuses on maximizing $\mathbb{E} \left( u_{e,l} \right)$ while the other policies are encouraged to choose different actions other than that chosen by policy $e$. Such iterative training will lead to $\lim\limits_{t\to\infty}\sigma_e(t) = 1$. Therefore, we note that NesPE in this context will converge to a pure strategy with a single policy $e$ that fully optimizes $\mathbb{E} \left( u_{e,l} \right)$. 

On the other hand, if there is no dominating action and a mixed strategy is preferred, $\sigma_e$ will be upper bounded. Each policy $e_1$ is encouraged to find a balance between optimizing $u_{e_1,l}(t)$ and diverge from others, while the better policies will be less affected by the correlation minimization. Thus, the player may obtain the optimized mixed strategy via the converged Nash equilibrium profile.

\subsection{Numerical Results} \label{subsec: equilexp}
In this section, we apply NesPE to both the  network slicing victim agent and the jammer agent, and compare the performances in terms of victims' average reward and completion ratio during the testing phase. All parameters other than the ensemble are the same as in Section \ref{subsec: jamexp}. 

For the policy ensemble of the victim user agent, we consider $E=5$ policies in the ensemble, and $L=5$ different classes of subsequent observations. In this case, at each base station $b$, each element $H^b[u, c]$ of the transmission rate history is a queue with the maximal length of $Q$ instead of a single number, so we have $H^b \in \mathbb{R}_{\ge 0}^{N_u \times N \times Q}$. The subsequent observation is the set of experienced transmission rate $\mathsf{r}_c^{b,u}(t)$ at time $t$, and is classified into type $l$ via
\begin{equation} \label{eq:latter_ob_user}
\small
\begin{aligned} 
l =& \operatorname*{argmin}_l \Big( d_o(H^b[u, c][Q-1]), d_o(H^b[u, c][Q-2]), \\
&d_o(H^b[u, c][Q-3]), d_o(H^b[u, c][Q-4]), d_o(\operatorname*{min}(H^b[u, c])) \Big),
\end{aligned} 
\end{equation}
\begin{equation} \label{eq:do}
\small
d_o(\mathsf{r}) = \sum_{u \in \mathcal{U}_b} \sum_{c \in C_u} \left( \mathsf{r} - \mathsf{r}_c^{b,u}(t) \right)^2,
\end{equation}
where $\mathcal{U}_b$ is the set of the served users at base station $b$, and $C_u$ is the set of channels in the slice assigned to user $u$.

For the policy ensemble of the jammer agent, we consider $E_J=2$ policies in the ensemble, and $L_J=2$ different classes of subsequent observations. In this case, a queue with a finite length of listened difference $d^\text{listen}_c$ is kept, and we use its average value $\overline{d^\text{listen}_c}$ to classify the subsequent observation $d^\text{listen}_c(t)$ into
\begin{equation} \label{eq:latter_ob_jam}
\begin{aligned} 
l_J & = \begin{cases}
1 \hspace{.3cm} \text{if $d^\text{listen}_c(t) < \overline{d^\text{listen}_c}$,}\\
2 \hspace{.3cm} \text{otherwise.}
\end{cases}\\
\end{aligned} 
\end{equation}

For comparison, we also show the results obtained with two other types of policy ensemble methods that aim at robustness, including agents with policy ensembles (APE) \cite{lowe2017multi} and cooperative evolutionary reinforcement learning (CERL) \cite{khadka2019collaborative}. The key idea of APE is to randomly initialize different policies, select one policy at a time, maintain a separate replay buffer, and maximize the expected reward. On the other hand, CERL sets different policies with different hyper-parameters, uses a shared replay buffer, and applies a neuro-evolutionary algorithm.

In Table \ref{tab:perform_pe}, we compare the victim agents' average rewards and completion ratios in the presence of a jammer agent when the proposed NesPE  and other algorithms are employed at both the network slicing agents and the jammer. In rows from top to bottom, we have the performance with no jammer, original jammer with a single policy, and NesPE, APE, and CERL based jammers, respectively. {For each jammer, we present the performance with both the lower power budget of 0dB where $P_J = P_B$, and the higher power budget of 60dB where $P_J = {10}^{6} P_B$.} Across different columns from left to right, we have the performance when the victim utilizes a single policy, and NesPE, APE, and CERL based policies, respectively. In the first row, we notice that each type of victim agent has a similar performance (in terms of both average reward and completion ratio/percentage), which indicates that different victim agents are able to identify and utilize the dominating strategy. In the following rows, we see that NesPE based victim network slicing agent achieves a better performance than the other victim agents against all different types of the jammer agent. In the third row in which the performance of the victim network slicing agents in the presence of a NesPE based jammer is provided, we observe that all victims under this NesPE based jamming attack have lower rewards and completion ratios compared to those under the other types of jamming attacks. {With the higher jamming power budget of 60dB, we observe similar trends when comparing different algorithms.} Hence, NesPE based jamming agent performs better in suppressing the performance of network slicing agents. This lets us conclude that NesPE based strategies outperform the other algorithms in a competitive environment where the adversary exists, and both the victim and jammer perform more favorably when they employ NesPE.

\savebox{\tempbox}{\begin{tabular}{@{}r@{}l@{\space}}
& Victim \\ Jammer
\end{tabular}}
\begin{table}[h!]
	\small
	\begin{center}
		\caption{Average Reward and Completion Ratio Comparison of Different Policy Ensemble Algorithms {and Different Jamming Power Budget} during Test Phase}
		\label{tab:perform_pe}
		\begin{tabular}{ | l || p{10mm} | p{10mm} |  p{10mm} | p{10mm} | }
			\hline
 			%\tikz[overlay]{\draw (0pt,\ht\tempbox) -- (\wd\tempbox,-\dp\tempbox);}
            \tikz[overlay]{\draw (-0.21,\ht\tempbox) -- (2.34,-\dp\tempbox);}
            \usebox{\tempbox}\hspace{\dimexpr 1pt-\tabcolsep}
            & Single & NesPE & APE & CERL  \\ \hline\hline
			None & 19.90 \newline 94.78\% & 20.08 \newline 95.23\% & 20.07 \newline 95.16\% & 20.08 \newline 95.22\%  \\ \hline
			Single, 0dB & 10.09 \newline 73.68\% & 14.96 \newline 84.90\% & 11.53 \newline 77.04\% & 11.71 \newline 77.30\%  \\ \hline
			NesPE, 0dB & 9.68 \newline 72.57\% & 12.22 \newline 78.76\% & 11.36 \newline 76.92\% & 11.23 \newline 76.31\%  \\ \hline
			APE, 0dB & 12.93 \newline 80.26\% & 14.44 \newline 83.55\% & 13.23 \newline 80.75\% & 11.42 \newline 76.66\%  \\ \hline
			CERL, 0dB & 12.38 \newline 78.81\% & 14.53 \newline 84.10\% & 12.53 \newline 79.20\% & 11.28 \newline 76.68\%  \\ \hline
			{Single, 60dB} & 3.58 \newline 58.76\% & 11.15 \newline 76.52\% & 8.58 \newline 70.33\% & 5.23 \newline 62.16\%  \\ \hline
			{NesPE, 60dB} & 2.25 \newline 55.35\% & 11.33 \newline 76.81\% & 6.69 \newline 66.19\% & 4.49 \newline 61.12\%  \\ \hline
			{APE, 60dB} & 3.35 \newline 58.09\% & 11.8 \newline 77.72\% & 9.44 \newline 72.39\% & 5.18 \newline 62.49\%  \\ \hline
			{CERL, 60dB} & 2.61 \newline 56.14\% & 11.98 \newline 78.15\% & 8.91 \newline 71.10\% & 4.70 \newline 61.65\%  \\ \hline
		\end{tabular}
	\end{center}
\end{table}

\section{Conclusion}\label{sec: con}
In this paper, we designed network slicing agents using MACC multi-agent deep RL with pointer network based actors. We considered an area covered by multiple base stations and addressed a dynamic environment with time-varying channel fading, mobile users, and randomly arriving service requests from the users. We described the system model, formulated the network slicing problem, and designed the MACC deep RL algorithm and pointer network based actor structure. We demonstrated the proposed agents' performance via simulations and have shown that they can achieve high average rewards and complete around $95\%$ of the requests.  

%by comparing against three statistical algorithms and three other deep RL based algorithms. The proposed network slicing agents are shown to outperform by completing 95\% of the requests and achieving the highest sum rate rewards under a demanding parameter setting.

Subsequently, we developed a deep RL based jammer that aims at minimizing the network slicing agents' (i.e., victims') transmission rate and thus degrades their performance without prior knowledge of the victim policies and without receiving any direct feedback. We introduced the jamming and listening phases of the proposed jammer and addressed the jamming location optimization. We also studied jamming channel optimization by designing an actor-critic agent that decides on which channels to jam. We have demonstrated the effectiveness of the proposed jammer via simulation results, and quantified the degradation in the performance of the network slicing agents compared to the performance achieved in the absence of any jamming attacks. We also provided comparisons among actor-critic based jammers with different assumptions on how to decide on which channels to jam (e.g., based on the last or estimated next interference or linear interpolation of the two).  

%Subsequently, we introduced a deep RL based jammer that aims at minimizing the  victim network slicing agents' transmission rate and thus degrades their performance without prior knowledge of the victim policies or direct feedback. We introduced its jamming phase and listening phase, the jamming location optimization, and the jamming channel optimization via actor-critic agent. We have demonstrated this jammer's effectiveness via experiences with the power budget much less than the base stations. 

Finally, we designed the NesPE algorithm for a competitive zero-sum game between the victim agents and jammer agent. By applying NesPE on both the victim agents and the jammer agent and comparing with two other policy ensemble algorithms, we have shown that NesPE not only adapts to a highly dynamic environment with time-varying fading and mobile users, but also adapts to a competitive environment against adversary's policy ensemble agent with optimal mixed strategy. Thus, both the victim network slicing agents and jammer should apply NesPE to attain improved performance levels, and the interaction between them converges to the Nash equilibrium over all possible policy ensembles.

\bibliographystyle{IEEEtran}
\bibliography{ref}

\begin{IEEEbiography}[{\includegraphics[width=1in,height=1.25in,clip,keepaspectratio]{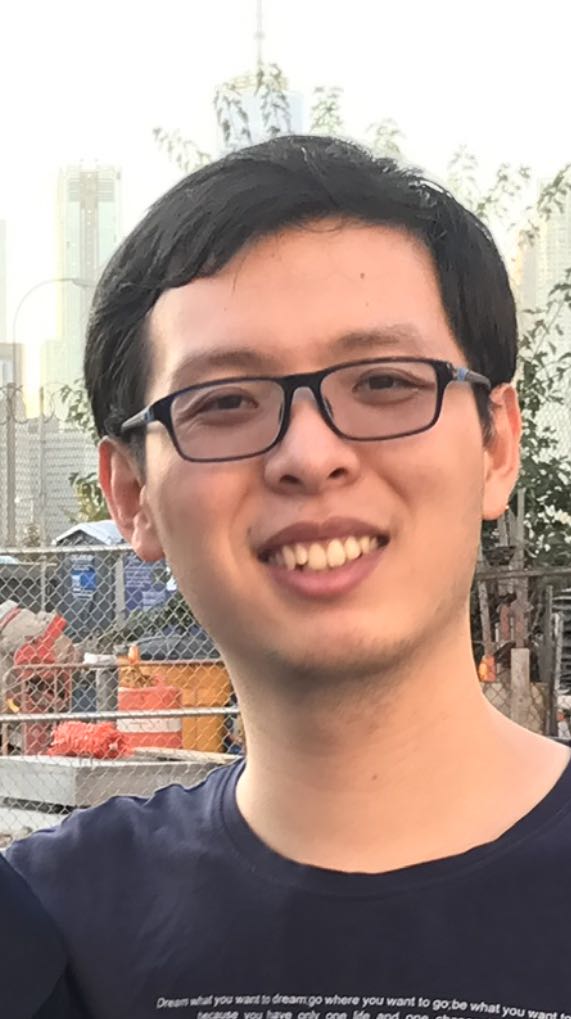}}]{Feng Wang} 
    is currently a Ph.D. student in the Department of Electrical Engineering and Computer Science at Syracuse University. He received his B.S. degree in Automation Science and Electrical Engineering from Beihang University (China) in 2016 and his M.S. degree from New York University in 2019. His primary research interests include wireless networks, adversarial learning, reinforcement learning, and federated learning.
\end{IEEEbiography}

\begin{IEEEbiography}[{\includegraphics[width=1in,height=1.25in,clip,keepaspectratio]{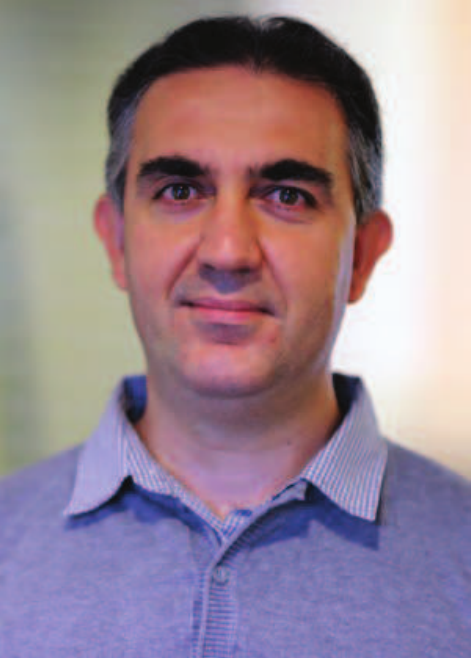}}]{M. Cenk Gursoy}
	received the B.S. degree with high distinction in electrical and electronics engineering from Bogazici University, Istanbul, Turkey, in 1999 and the Ph.D. degree in electrical engineering from Princeton University, Princeton, NJ, in 2004. He was a recipient of the Gordon Wu Graduate Fellowship from Princeton University between 1999 and 2003. He is currently a Professor in the Department of Electrical Engineering and Computer Science at Syracuse University. His research interests are in the general areas of wireless communications, information theory, communication networks, signal processing, optimization and machine learning. He is an Editor for IEEE Transactions on Communications, and an Area Editor for IEEE Transactions on Vehicular Technology. He is on the Executive Editorial Committee of IEEE Transactions on Wireless Communications.  He also served as an Editor for IEEE Transactions on Green Communications and Networking between 2016--2021, IEEE Transactions on Wireless Communications between 2010--2015 and 2017--2022, IEEE Communications Letters between 2012--2014, IEEE Journal on Selected Areas in Communications - Series on Green Communications and Networking (JSAC-SGCN) between 2015-2016, Physical Communication (Elsevier) between 2010--2017, and IEEE Transactions on Communications between  2013--2018. He has been the co-chair of the 2017 International Conference on Computing, Networking and Communications (ICNC) - Communication QoS and System Modeling Symposium, the co-chair of 2019 IEEE Global Communications Conference (Globecom) - Wireless Communications Symposium, the co-chair of 2019 IEEE Vehicular Technology Conference Fall - Green Communications and Networks Track, and the co-chair of 2021 IEEE Global Communications Conference (Globecom), Signal Processing for Communications Symposium.  He received an NSF CAREER Award in 2006. More recently, he received the EURASIP Journal of Wireless Communications and Networking Best Paper Award, 2020 IEEE Region 1 Technological Innovation (Academic) Award,  2019 The 38th AIAA/IEEE Digital Avionics Systems Conference Best of Session (UTM-4) Award, 2017 IEEE PIMRC Best Paper Award, 2017 IEEE Green Communications \& Computing Technical Committee Best Journal Paper Award, UNL College Distinguished Teaching Award, and the Maude Hammond Fling Faculty Research Fellowship. He is a Senior Member of IEEE, and is the Aerospace/Communications/Signal Processing Chapter Co-Chair of IEEE Syracuse Section.
\end{IEEEbiography}

\begin{IEEEbiography}[{\includegraphics[width=1in,height=1.25in,clip,keepaspectratio]{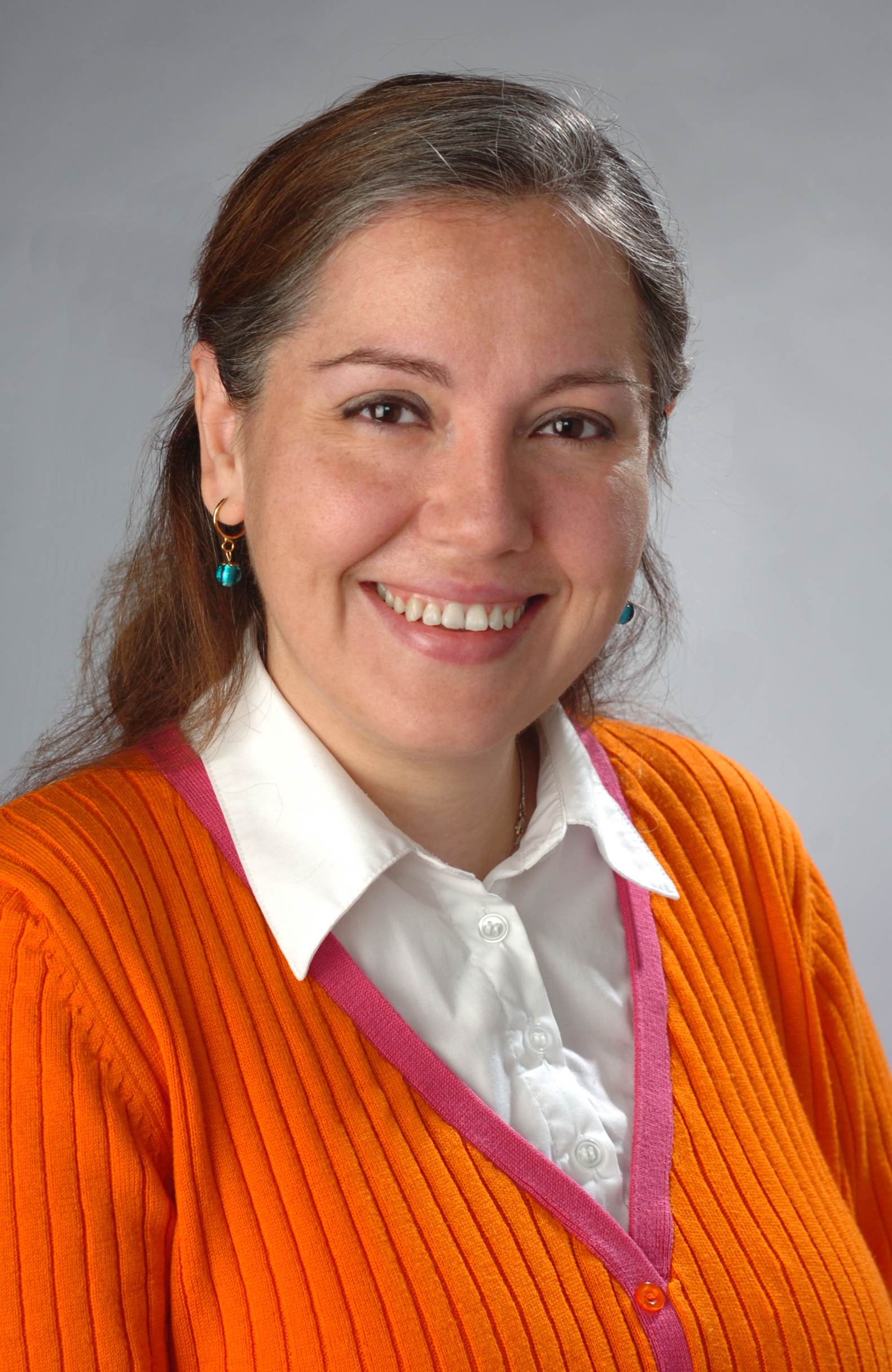}}]{Senem Velipasalar}
	(Senior Member, IEEE) received the B.S. degree in electrical and electronic engineering from Bogazici University, Istanbul, Turkey, in 1999, the M.S. degree in electrical sciences and computer engineering from Brown University, Providence, RI, USA, in 2001, and the M.A. and Ph.D. degrees in electrical engineering from Princeton University, Princeton, NJ, USA, in 2004 and 2007, respectively. From 2007 to 2011, she was an Assistant Professor with the Department of Electrical Engineering, University of Nebraska-Lincoln. She is currently a Professor with the Department of Electrical Engineering and Computer Science, Syracuse University. The focus of her research has been on machine learning and its applications to 3D point cloud analysis, occupancy detection, and video activity recognition, UAV-mounted and wearable camera applications, multicamera tracking, and surveillance systems. She is a member of the Editorial Board of the IEEE Transactions on Image Processing and Journal of Signal Processing Systems (Springer).
\end{IEEEbiography}

\end{document}